\documentclass[runningheads]{llncs}

\usepackage{eccv}

\usepackage{eccvabbrv}

\usepackage{graphicx}
\usepackage{booktabs}

\usepackage[accsupp]{axessibility}  % Improves PDF readability for those with disabilities.

\usepackage{hyperref}

\usepackage{orcidlink}

\usepackage{amsmath,amsfonts,bm}

\def\eqref#1{equation~\ref{#1}}
\def\1{\bm{1}}
\DeclareMathAlphabet{\mathsfit}{\encodingdefault}{\sfdefault}{m}{sl}
\SetMathAlphabet{\mathsfit}{bold}{\encodingdefault}{\sfdefault}{bx}{n}
\usepackage[utf8]{inputenc}
\usepackage[T1]{fontenc}
\usepackage{amsmath, amssymb, amsfonts}
\usepackage{mathtools}        % loads/extends amsmath
\usepackage{bm}
\usepackage{mathrsfs}
\usepackage{bbm}

\usepackage{booktabs}
\usepackage{tabularx}
\usepackage{array}
\usepackage{makecell}
\usepackage{multirow}
\usepackage{multicol}
\usepackage{colortbl}
\usepackage{hhline}
\usepackage{diagbox}
\usepackage{enumitem}
\usepackage{placeins}
\usepackage{arydshln}

\usepackage{graphicx}
\usepackage{wrapfig}
\usepackage{tikz}
\usepackage{pgfplots}
\pgfplotsset{compat=1.17}
\usepackage[inkscapearea=page]{svg} % or inkscapearea=drawing
\usepackage{fontawesome}

\usepackage{algorithm, algorithmic}

\usepackage{microtype}
\usepackage{xurl}      % better URL line breaks (supersedes bare url)
\usepackage{xspace}
\usepackage{alltt}
\usepackage{nicefrac}
\usepackage{scalerel}
\usepackage{marginnote}
\usepackage{outlines}
\usepackage{capt-of}
\usepackage{soul}
\usepackage{lipsum}    % dummy text (optional; remove in production)

\usepackage[dvipsnames,table]{xcolor}
\definecolor{Gray}{gray}{0.9}
\definecolor{lightgray}{gray}{0.95}
\definecolor{cvprblue}{rgb}{0.21,0.49,0.74}
\definecolor{coco1}{HTML}{D9E4EC}
\definecolor{coco2}{HTML}{B7CFDC}
\definecolor{coco3}{HTML}{6AABD2}
\definecolor{coco4}{HTML}{385E72}

\usepackage{tcolorbox}
\tcbset{
  mycallout/.style={
    colback=gray!5,
    colframe=black!70,
    boxrule=0.8pt,
    arc=2pt,
    left=8pt,right=8pt,top=6pt,bottom=6pt,
  }
}

\usepackage{float}

\usepackage{pifont}
\usepackage{savesym}

\usepackage{dingbat}
\usepackage[misc]{ifsym}
\newcommand{\cmark}{\Large{\textcolor{ForestGreen}{\ding{51}}}} % ✓ (from pifont)
\newcommand{\xmark}{\textcolor{Magenta}{\ding{55}}} % ✗
\usepackage{tikz}
\usetikzlibrary{positioning}

\usepackage{tikz}

\newcolumntype{Y}{>{\raggedright\arraybackslash}X}
\usepackage{calc}

\usepackage[capitalize,noabbrev]{cleveref}

\usepackage{rotating}

\newcommand{\linner}[2]{\left\langle #1,\, #2 \right\rangle_{\mathcal{L}}}

\begin{document}

% ---------------------------------------------------------------
% \title{Hyperbolic Geometric Alignment\\for Vision-Language Dataset Distillation}
% \title{Geometric Range–Null Distillation in Hyperbolic Space for Image–Text Learning}
% \title{
% \includesvg[width=17pt,graphicsopts={trim=20mm 20mm 20mm 80mm, clip}]{figs/matcha_leaf}\;%
% MATCHA: Matching Hyperbolically Aligned Distributions for Vision–Language Dataset Distillation
% }
\title{Rank-Aware Hyperbolic Alignment \\for Vision–Language Dataset Distillation}

\titlerunning{RAHA}

\author{
Jongoh Jeong~\inst{1}~\orcidlink{0000-0002-5354-2693},
Sun-Kyung Lee~\inst{2}~\orcidlink{0000-0002-6566-8509}, 
\and
Kuk-Jin Yoon~\inst{1}~\orcidlink{0000-0002-1634-2756} 
}

\authorrunning{Jeong \etal}

\institute{Korea Advanced Institute of Science and Technology (KAIST), Republic of Korea \and
Electronics and Telecommunications Research Institute (ETRI), Republic of Korea\\
\email{\{jeong2, kjyoon\}@kaist.ac.kr, sklee2014@etri.re.kr}
\;\;\href{https://andyj1.github.io/raha}{\textcolor{black}{\small \faGithub\;Project Page}}
}

\maketitle

\begin{abstract}

    Vision-language dataset distillation (VLDD) compresses a large image-text paired dataset into a small set of synthetic pairs that can efficiently train contrastive vision-language models under strict data and compute budgets. Most existing methods match expert trajectories or cross-modal statistics, yet still enforce full-dimensional alignment in a Euclidean embedding space. This is often overly restrictive due to rank-deficient image--text correlation, with shared semantics concentrated in a low-dimensional range and remaining variation spread across a weakly correlated residual subspace. LoRS relaxes alignment at the similarity level by low-rank factorization, but does not explicitly control dominant alignment capacity and structure in the representation space. We thus propose a rank-aware hyperbolic alignment (RAHA) that combines hierarchical geometry with explicit alignment-capacity control. RAHA lifts multimodal representations to hyperbolic space and optimizes distilled pairs with asymmetric objectives that enforce geodesic alignment in the shared range while regularizing the residual subspace to preserve modality-private diversity and improve transfer robustness. Experiments on benchmarks show that RAHA demonstrates competitive cross-modal retrieval and improved transfer indicators under fixed budgets.

\end{abstract}
\section{Introduction}
\label{sec:intro}

Vision--language models (VLMs) have become a foundational layer of modern AI, connecting perception with language to support retrieval, captioning, grounding, and multimodal reasoning~\cite{blip, siglip, flamingo, llava, qwen, gpt4, gemini1_5}. Their rapid progress is driven by large paired image--text datasets and scalable contrastive objectives, amplified by careful data mixture design, filtering, deduplication, and hard-negative effects in batch training. As paired collections grow from millions to billions of pairs~\cite{kakaobrain2022coyo-700m, laion-5b, yfcc100m, COCO, hodosh2013flickr8k, young2014flickr}, capability improves, but the associated burdens escalate. These burdens include privacy exposure from unconsented personal data~\cite{thiel2023csam, birhane2023laionhate}, uncertain licensing across countries and platforms~\cite{longpre2023provenance}, limited provenance traceability and consent management~\cite{longpre2024consent}, and vulnerability to data poisoning and content drift~\cite{carlini2023poisoning}. These constraints increasingly shape how multimodal models can be built, audited, shared, and deployed.

This tension motivates a practical question: \emph{can we replace web-scale paired data with compact and auditable surrogates that retain multimodal training utility?} Dataset distillation (DD) offers a principled direction by synthesizing a small set of examples that approximates the learning signal of a much larger dataset~\cite{wang2018datasetdistillation, lei2023ddsurvey1, yu2023ddsurvey2, dd_survey25}. Beyond efficiency, distilled surrogates can reduce the footprint of sensitive data, enable controlled redistribution when raw pairs cannot be shared, and accelerate research iteration through cheaper ablations and faster model selection. In this sense, distillation is not only a systems optimization, but also a practical tool for developing VLMs under real constraints.

Extending DD to vision--language data is harder than the unimodal case as a distilled set must preserve \emph{cross-modal relational structure}, not only unimodal diversity. Table~\ref{tab:mdd_taxonomy_singlecol} organizes prior VLDD methods by the supervision signal extracted from real data and the spaces in which they distill images and text. Existing approaches broadly fall into three families: (i) Trajectory-based methods~\cite{wu2024vldistill, xu2024lors, zhang2025repblend} match training dynamics but incur high cost and can inherit architectural bias from the teacher; (ii) Generative methods~\cite{zhao2025edge} leverage diffusion priors to synthesize pairs, trading direct control of alignment structure for scalability; and (iii) Distribution statistics-based methods~\cite{lee2025covmatch} match cross-modal moments efficiently, yet they operate in the Euclidean space and still impose a largely uniform alignment constraint across feature directions. Across these families, a common limitation persists: \emph{current methods provide limited explicit control over which representation directions should be aligned and which should remain modality-private, especially under the extreme compression.}
%regime of VLDD.}

This limitation is consequential because image--text correlation is often effectively \emph{low-rank}~\cite{xu2024lors}. A compact shared subspace captures dominant semantic factors and coarse relations that should align across modalities, while the remaining directions form a weakly correlated \emph{residual} component that absorbs modality-specific cues, annotation artifacts, and nuisance variation. Under tight budgets, enforcing alignment uniformly in this residual component can suppress complementary information and harm transfer. Separately, multimodal semantics are naturally hierarchical, spanning entities, attributes, and relations at different abstraction levels. Euclidean geometry offers limited inductive bias for preserving such nested structure when only a small distilled set must carry the training signal.

We thus address these limitations altogether by proposing \textbf{RAHA} (\textbf{R}ank-\textbf{A}ware \textbf{H}yperbolic \textbf{A}lignment), a geometry-aware distillation framework that controls \emph{what} is aligned and \emph{how}. As shown in Table~\ref{tab:mdd_taxonomy_singlecol}, RAHA couples hyperbolic contrastive alignment with an explicit range--residual treatment of cross-modal correlation. RAHA estimates an adaptive rank decomposition of real batch coupling, matches synthetic relevance in the shared \emph{range} component, and regularizes the complementary \emph{residual} component so that weak interactions do not dominate under compression. These objectives complement a base hyperbolic contrastive term that maintains pairwise discriminability on the synthetic set. We evaluate RAHA on standard VLDD benchmarks and analyze retrieval, cross-architecture transfer, robustness to perturbations, and qualitative fidelity of synthesized pairs, with the goal of understanding when structured relevance distillation is most beneficial.

\noindent \textbf{Contributions.} We make the following main contributions in this paper:
\vspace{-2mm}
\begin{itemize}[leftmargin=*]\setlength\itemsep{0.25em}
\item \textbf{Rank-aware hyperbolic formulation for VLDD.}\quad We formulate VLDD as hyperbolic contrastive learning with tangent-space relevance distillation that exposes an explicit range--residual decomposition of cross-modal correlation.
\item \textbf{Range--residual relevance distillation objective.}\quad We propose an objective that matches real and synthetic relevance in the shared dominant range component while regularizing the residual component to preserve modality-private variability under tight budgets.
\item \textbf{Evaluation with component ablations.}\quad We evaluate RAHA on standard retrieval benchmarks and provide ablations that isolate the contributions of hyperbolic contrast, range matching, and residual regularization, alongside analyses of cross-architecture transfer, robustness, and qualitative samples.
\end{itemize}
\vspace{-3pt}

\begin{table}[!t]
    \centering
    \caption{\textbf{Comparison in the approach,} categorized by source of supervision from real data, latent, image/text spaces each method operates on with the respective focus.}
    \label{tab:mdd_taxonomy_singlecol}
    \vspace{-3mm}
    
    \renewcommand{\arraystretch}{1.15}
    \renewcommand{\aboverulesep}{2pt}
    \renewcommand{\belowrulesep}{2pt}
    \setlength{\tabcolsep}{10pt}
    \scriptsize
    \fontsize{18}{18}\selectfont % 24 on 36(baseline skip)
    
    \resizebox{\linewidth}{!}{%
    \begin{tabular}{lcccc|l}
    \toprule
        \textbf{Method} &
        \textbf{Source of supervision} &
        \textbf{Latent space} &
        \textbf{Image space} &
        \textbf{Text space} &
        \multicolumn{1}{c}{\textbf{Approach/Focus}} \\
        % \textbf{Remark} \\
    \midrule

        MTT-VL~\cite{wu2024vldistill} &
        \multirow{3}{*}{Training Trajectory (Expert)} &
        \multirow{5}{*}{Euclidean} &
        Pixel &
        Enc. output &
        Matching Training Trajectory \\
        % bi-trajectory; frozen text \\
    \cmidrule(lr){1-1}\cmidrule(lr){4-6}

        LoRS~\cite{xu2024lors} &
        % Training Trajectory (Expert) &
        &&
        Pixel &
        Enc. output &
        + Low-rank similarity \\
        % relevance mining; efficient negatives \\
    \cmidrule(lr){1-1}\cmidrule(lr){4-6}

        RepBlend~\cite{zhang2025repblend} &
        % Training Trajectory (Expert) &
        &&
        Pixel &
        Enc. output &
        + Modality collapse mitigation \\
        % representation blending; anti-collapse \\
    \cmidrule(lr){1-2}\cmidrule(lr){4-6}

        EDGE~\cite{zhao2025edge} &
        Diffusion Model (SD~\cite{rombach2022SD}) &
        &
        DM latent &
        Caption &
        Generative \\
        % diffusion prior; caption augment \\
    \cmidrule(lr){1-2}\cmidrule(lr){4-6}

        CovMatch~\cite{lee2025covmatch} &
        Distribution (Statistics) &     
        &
        Pixel &
        Enc. input &
        Image-text cross-covariance \\
        % cross-covariance; trainable text \\
    \midrule
    \rowcolor{eccvblue!20}
        \textbf{Ours} &
        Distribution (Statistics) &
        Hyperbolic &
        Pixel &
        Enc. input &
        Explicit subspace decomposition \\
        % Remark \\
        
    \bottomrule
    \end{tabular}%
    }
\end{table}

\section{Related Work}\label{sec:relatedwork}
\noindent\textbf{Unimodal dataset distillation} (DD)
synthesizes a small surrogate set that preserves the training utility of a much larger corpus~\cite{wang2018datasetdistillation}. Early kernel-based formulations~\cite{KIP,KIP2} offer analytical insight but scale poorly. More practical approaches match training signals or feature statistics: gradient matching~\cite{DC,DSA}, distribution matching~\cite{CAFE,DM,IDM}, and efficient parameterizations of synthetic data~\cite{IDC,haba,KFS,RFAD,FRePo}. In parallel, trajectory matching aligns full optimization paths~\cite{MTT} and has been extended with memory-efficient scaling~\cite{cui2023tesla} and improved stability~\cite{du2023ftd,ATT,guo2023datm,zhong2025mct}. Recent work broadens the toolkit further with implicit-gradient views~\cite{RCIG}, information-theoretic criteria~\cite{shang2023mim4dd}, optimal-transport objectives~\cite{liu2025wmdd,cui2025optical}, and diffusion-based generation~\cite{Generative,su2024d4m,wang2025cao2}. We further refer to~\cite{lei2023ddsurvey1,yu2023ddsurvey2,cui2022dcbench} for comprehensive surveys. These unimodal advances provide the methodological foundation, but the field is transitioning to the multimodal regime, where distilled sets must additionally preserve cross-modal correspondence.% under contrastive objectives.

\noindent\textbf{Vision--language dataset distillation} (VLDD)
extends DD to paired image--text data, where a compact synthetic set must maintain both intra-modal diversity and cross-modal alignment. MTT-VL~\cite{wu2024vldistill} initiated the field with trajectory matching and retrieval-based evaluation. LoRS~\cite{xu2024lors} improves efficiency by distilling low-rank similarity structure within the same paradigm. RepBlend~\cite{zhang2025repblend} addresses multimodal-specific pathologies such as modality collapse through representation blending and balanced supervision. Moving beyond expert trajectories, distribution matching approaches align cross-modal statistics via geodesic kernel energy on a unit hypersphere~\cite{jeong2026mdm} and cross-covariance~\cite{lee2025covmatch} with within-modality regularization, where \cite{lee2025covmatch} is the first to train the text encoder jointly. In contrast, EDGE~\cite{zhao2025edge} leverages diffusion priors to generate pairs with correlation and diversity objectives. While these existing methods operate in Euclidean space with uniform alignment pressure, our method is trajectory-free and geometry-aware: we perform \emph{rank-aware hyperbolic alignment} that decomposes cross-modal correlation into a shared range and a residual subspace, selectively enforcing alignment only where it carries semantic content while preserving modality-private diversity.

\noindent\textbf{Hyperbolic geometry for vision--language.}\quad
Hyperbolic spaces are Riemannian manifolds with constant negative curvature that embed tree-like hierarchies with low distortion~\cite{nickel2018lorentz,peng2021hyperbolic_survey}. MERU~\cite{desai2023meru} maps image--text features onto the Lorentz model and reveals an interpretable radial layout in which generic concepts lie near the origin, suggesting that the modality gap can reflect abstraction-level differences rather than mere misalignment. Ramasinghe~\emph{et al.}~\cite{ramasinghe2024acceptgap} argue that alignment objectives should respect, not collapse, unimodal hierarchies. On the data-centric side, HDD~\cite{li2025hdd} studies hyperbolic centroid matching for unimodal distillation, and recent hyperbolic VL methods exploit entailment-like asymmetry through hierarchy-aware objectives~\cite{pal2025hycoclip,kim2024hype,poppi2025hysac}. We build on these foundations but introduce an explicit \emph{range--residual decomposition} within hyperbolic space, enabling selective alignment along shared semantic directions while suppressing information-deficient residual components to preserve hierarchical cross-modal structure under extreme compression.

\section{Proposed Method}
\label{sec:method}

We propose \textbf{Rank-Aware Hyperbolic Alignment} (RAHA), a vision--language dataset distillation method that learns a compact synthetic set by optimizing two objectives jointly: (i) a \emph{hyperbolic contrastive} loss on synthetic image--text pairs, and (ii) a \emph{rank-aware relevance distillation} loss that transfers cross-modal rank structure from real to synthetic data through a range--residual decomposition of batch-level statistics, as highlighted in Table~\ref{tab:distinction}.

\begin{table}[t]
  \centering
  \captionsetup{font={small,stretch=1.15}}
  \setlength{\tabcolsep}{4pt}
  \renewcommand{\arraystretch}{1.12}
  \caption{\textbf{Comparison of structural information preserved and missed in the existing methods.}
    Left: matched schematic targets. Right: description of targeted structures versus our proposed rank-aware hyperbolic alignment.
  \scriptsize{$^{\ast}$Not in particular order.}}
  \label{tab:distinction}
  \vspace{1mm}

  \begin{minipage}[t]{0.43\linewidth}
    \centering
    \vspace{2mm}
    \begin{minipage}[c][4.2cm][c]{0.96\linewidth}
      \centering
      \vspace{-2mm}
      \includegraphics[width=\linewidth]{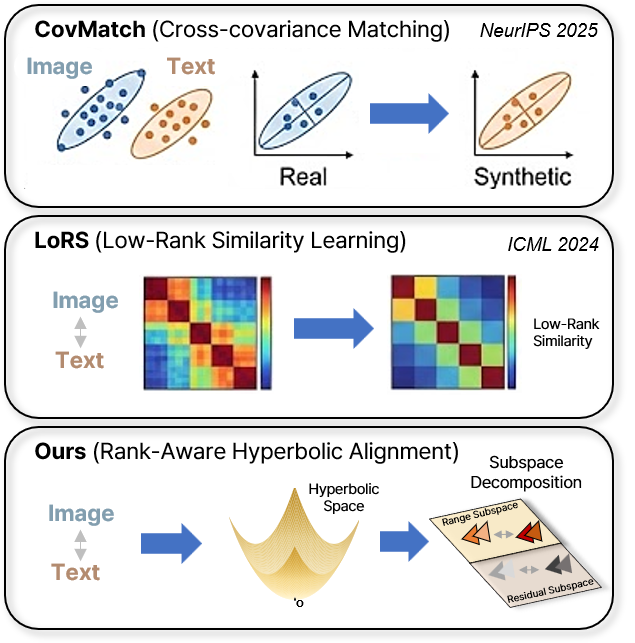}\\[-2mm]
      {\scriptsize (a) Matching target structures}
    \end{minipage}
  \end{minipage}\hfill
  \begin{minipage}[t]{0.57\linewidth}
    \centering
    % \vspace{-7mm}
    % \caption{Matching target by method.}
    \vspace{-7mm}
    \setlength{\tabcolsep}{2pt}
    \renewcommand{\aboverulesep}{0.7pt}
    \renewcommand{\belowrulesep}{0.7pt}

    \resizebox{\linewidth}{!}{%
      \begin{tabular}{
          >{\centering\arraybackslash}m{1.9cm}
          >{\raggedright\arraybackslash}m{2.8cm}
          >{\raggedright\arraybackslash}p{3.5cm}
        }
        \multicolumn{3}{c}{(b) Matching target by method}\\
        \toprule
        \textbf{Method} & \textbf{Matched Target} & \textbf{Structure} \\
        \midrule

        CovMatch~\cite{lee2025covmatch} &
        Cross-covariance \quad(+ per-modality feature reg.) &
        \vspace{-7mm}\textbullet~\textbf{Preserves:} 2$^{nd}$-order cross-modal correlation\newline
        \textbullet~\textbf{Misses:} Explicit pairwise ranking / higher-order structure \\
        \midrule

        LoRS~\cite{xu2024lors} &
        Explicit similarity matrix $\tilde{S}$ with low-rank factorization &
        \vspace{-9mm}\textbullet~\textbf{Preserves:} Pairwise similarity via low-rank structure\newline
        \textbullet~\textbf{Misses:} Geometry-/hierarchy-aware structure \\
        \midrule

        Ours &
        Rank-aware image-text relevance by subspace &

        \vspace{-7mm}\textbullet~\textbf{Preserves:} Relative similarity \emph{order} + hierarchical structure
        % \textbullet~\textbf{Misses:} --
        \\
        \bottomrule
      \end{tabular}%
    }
  \end{minipage}

  \vspace{-5mm}
\end{table}

\subsection{Preliminaries and Distillation Protocol}
\label{sec:prelim}

\noindent\textbf{Scope and protocol.}\quad
Dataset distillation seeks a small synthetic set that preserves the training utility of a much larger dataset~\cite{wang2018datasetdistillation,DM,MTT}.
In vision--language learning, paired datasets are large and training must preserve cross-modal alignment.
Since retrieval-oriented VLMs use contrastive objectives, the distilled set must preserve \emph{bidirectional} ranking behavior (image-to-text and text-to-image).
VLDD is evaluated by training a retrieval model on the synthetic set and reporting Recall@K in both directions.

\smallskip
\noindent\textbf{Vision--language dataset distillation.}\quad
Let the real paired dataset be
$\mathcal{D}_{\mathrm{real}}=\{(x_i,t_i)\}_{i=1}^{N}$,
where $x_i\in\mathcal{X}$ is an image and $t_i\in\mathcal{T}$ is a caption.
We synthesize a compact set
$\mathcal{D}_{\mathrm{syn}}=\{(\tilde{x}_j,\tilde{t}_j)\}_{j=1}^{M}$ with $M\ll N$
such that training on $\mathcal{D}_{\mathrm{syn}}$ yields comparable bidirectional retrieval to training on $\mathcal{D}_{\mathrm{real}}$.
The retrieval model is
$\Psi=\{E_v(\cdot;\theta_v),\,E_t(\cdot;\theta_t),\,\pi_v(\cdot;\omega_v),\,\pi_t(\cdot;\omega_t)\}$,
where $E_v,E_t$ are modality-specific encoders and $\pi_v,\pi_t$ are linear projection heads that map encoder outputs to a shared $d$-dimensional space.
For a pair $(x,t)$, the projected embeddings are
\begin{equation}
  z^v=\pi_v(E_v(x;\theta_v);\omega_v)\in\mathbb{R}^d,\qquad
  z^t=\pi_t(E_t(t;\theta_t);\omega_t)\in\mathbb{R}^d.
  \label{eq:z_def}
\end{equation}
RAHA operates on these projected features without cosine normalization.

\smallskip
\noindent\textbf{Synthetic data parameterization.}\quad
Each synthetic image is a learnable tensor $\tilde{x}_j\in\mathbb{R}^{3\times H\times W}$.
Discrete text tokens are not differentiable, so we follow \cite{lee2025covmatch} to parameterize each synthetic caption by its input-layer token embeddings and an attention mask:
\begin{equation}
  \tilde{t}_j \equiv (\tilde{\mathbf{E}}_j,\tilde{\mathbf{m}}_j),\qquad
  \textrm{where}\;\;
  \tilde{\mathbf{E}}_j\in\mathbb{R}^{L\times d_e},\;\;
  \tilde{\mathbf{m}}_j\in\{0,1\}^{L},
  \label{eq:syn_text_param}
\end{equation}
where $L$ is the maximum sequence length and $d_e$ is the token embedding dimension.
The mask $\tilde{\mathbf{m}}_j$ is fixed at initialization, only marking valid-token/padding positions, while $\tilde{\mathbf E}_j$ carries learnable linguistic content.
This parameterization keeps $\mathcal{D}_{\mathrm{syn}}$ differentiable with respect to both image pixels and text embeddings.

\subsection{Hyperbolic Contrastive Alignment}
\label{sec:hyperbolic_hitc}

\noindent\textbf{Transition from Euclidean to hyperbolic InfoNCE.}\quad
Standard contrastive alignment uses InfoNCE with a Euclidean similarity (dot product or cosine) to push matched pairs together and mismatched pairs apart within a batch~\cite{radford2021clip}.
Given a batch of paired embeddings $\{(z_i^v,z_i^t)\}_{i=1}^{\tilde{B}}$, Euclidean image--text contrastive loss constructs logits $\ell_{ij}=s(z_i^v,z_j^t)/\tau$ with temperature $\tau>0$ and similarity function $s$, then applies symmetric cross-entropy over both directions.
RAHA replaces the Euclidean similarity with a geodesic distance on the Lorentz hyperboloid.
We base our intuition from the fact that cross-modal semantics that exhibit a hierarchical structure by which captions refine coarse visual concepts into specific attributes, and the hyperbolic geometry accommodates such nested relations through its exponential geometry~\cite{nickel2017poincare,nickel2018lorentz,ganea2018hyperbolic}.

\smallskip
\noindent\textbf{Lorentz model and lifting.}\quad
Let $c>0$ denote the curvature parameter.
A point on the Lorentz hyperboloid is $u=(u_0,\bar{u})\in\mathbb{R}^{d+1}$ satisfying
$-u_0^2+\|\bar{u}\|_2^2=-1/c$ with $u_0>0$.
Given a tangent vector $w\in\mathbb{R}^d$ at the origin with norm $r=\|w\|_2$, the exponential map produces the spatial component
\vspace{-2mm}
\begin{equation}
  \mathrm{Exp}_o^c(w)
  =
  \frac{\sinh(\sqrt{c}\,r)}{\sqrt{c}\,r}\,w \in\mathbb{R}^d,
  \label{eq:exp_space}
\end{equation}
and the time coordinate is recovered as $u_0=\sqrt{1/c+\|\mathrm{Exp}_o^c(w)\|_2^2}$.
Given a scale parameter $s>0$, we lift projected embeddings from Eq.~\ref{eq:z_def} to the hyperboloid:
\vspace{-2mm}
\begin{equation}
  h^v=\mathrm{Exp}_o^c(s\, z^v),\qquad h^t=\mathrm{Exp}_o^c(s\, z^t).
  \label{eq:lift}
\end{equation}
The geodesic distance between two points $u=(u_0,\bar{u})$ and $v=(v_0,\bar{v})$ on the hyperboloid is
\vspace{-2mm}
\begin{equation}
  d_c(u,v)=\frac{1}{\sqrt{c}}\operatorname{arcosh}\!\bigl(-c\,\langle u,v\rangle_{\mathcal{L}}\bigr),
  \label{eq:lorentz_dist}
\end{equation}
where $\langle u,v\rangle_{\mathcal{L}}=-u_0 v_0+\bar{u}^\top \bar{v}$ is the Lorentzian inner product.

\smallskip
\noindent\textbf{Hyperbolic image--text contrastive loss (hITC).}\quad
For a synthetic batch of size $\tilde{B}$, we define logits as the negative geodesic distance scaled by temperature $\tau>0$ (\ie, 0.07):
\vspace{-2mm}
\begin{equation}
  \ell_{ij}=-\frac{d_c(h_i^v,h_j^t)}{\tau}.
  \label{eq:hyp_logits}
\end{equation}
The hITC loss applies symmetric cross-entropy:
\vspace{-2mm}
\begin{equation}
  \mathcal{L}_{\mathrm{hITC}}
  =
  \frac{1}{2\tilde{B}}\sum_{i=1}^{\tilde{B}}
  \left[
    -\log \frac{\exp(\ell_{ii})}{\sum_{j}\exp(\ell_{ij})}
    -\log \frac{\exp(\ell_{ii})}{\sum_{j}\exp(\ell_{ji})}
  \right].
  \label{eq:hITC}
\end{equation}
This loss is computed on \emph{synthetic pairs only} and does not involve real data.

\subsection{Distilling Cross-Relevance via Range and Residual Subspaces}
\label{sec:raha}

$\mathcal{L}_{\mathrm{hITC}}$ enforces alignment within each synthetic pair but does not transfer the \emph{relative} cross-modal ranking structure observed in real data.
To distill this cross-modal structure, RAHA decomposes the cross-covariance of real tangent-space features into a low-rank \emph{range} subspace, capturing the adaptive top-\textit{k} dominant image--text coupling directions, and a complementary \emph{residual} subspace containing the remaining interactions.
Relevance distributions in each subspace are then matched from real to synthetic via entropy-regularized optimal transport with Sinkhorn iterations. The final loss groups each subspace's matching term with its regularizer, yielding a clean three-term objective: hITC, range, and residual.

\noindent\textbf{Tangent-space decomposition.}\quad
\label{sec:tangent_decomp}
%
% \noindent\textbf{Tangent features.}\quad
Both real and synthetic projected features are lifted to the hyperboloid via Eq.~\ref{eq:exp_space} and mapped back to the tangent space at the origin with the logarithmic map:
\begin{equation}
  x^v = \mathrm{Log}_o^c(h^v), \qquad x^t = \mathrm{Log}_o^c(h^t).
  \label{eq:log_map}
\end{equation}
This round-trip (Euclidean $\to$ hyperboloid $\to$ tangent space) applies a nonlinear warping controlled by $c$ and $s$ that concentrates features according to their hyperbolic norm, so that subsequent linear operations respect the underlying geometry.

\smallskip
\noindent\textbf{Hyperbolic tangent-space cross-covariance.}\quad
For a real batch $\mathcal{B}$ of size $B$, let $X^v,X^t\in\mathbb{R}^{B\times d}$ stack the tangent features row-wise.
Define batch means $\mu_v=\frac{1}{B}\sum_i x_i^v$, $\mu_t=\frac{1}{B}\sum_i x_i^t$ and the cross-covariance
\begin{equation}
  C_{\mathrm{real}}
  =
  \frac{1}{B-1}
  (X^v-\mathbf{1}\mu_v^\top)^\top
  (X^t-\mathbf{1}\mu_t^\top)\;\in\mathbb{R}^{d\times d}.
  \label{eq:xcov_real}
\end{equation}
$C_{\mathrm{syn}}\in\mathbb{R}^{d\times d}$ is computed identically from the synthetic batch.
Real features are treated as stop-gradient targets throughout.

\smallskip
\noindent\textbf{Range--residual subspace decomposition.}\quad
We factorize $C_{\mathrm{real}}$ via SVD:
\begin{equation}
  C_{\mathrm{real}}=U\Sigma V^\top,\qquad
  \Sigma=\mathrm{diag}(\sigma_1,\dots,\sigma_d),\quad \sigma_1\ge \cdots \ge \sigma_d\ge 0,
  \label{eq:svd_real}
\end{equation}
where each singular value $\sigma_i$ measures the strength of cross-modal coupling along that direction.
We select the effective rank $k$ as the smallest integer whose top-$k$ singular values capture at least a fraction $\rho\in(0,1)$ of the total squared energy:
\begin{equation}
  k=\min\!\left\{k'\!:\ \frac{\sum_{i=1}^{k'}\sigma_i^2}{\sum_{i=1}^{d}\sigma_i^2}\ge \rho\right\}.
  \label{eq:rank_energy}
\end{equation}
Let $U_k\in\mathbb{R}^{d\times k}$ and $V_k\in\mathbb{R}^{d\times k}$ denote the first $k$ columns of $U$ and $V$.
These define the range basis: $U_k$ spans the image-side coupling directions and $V_k$ spans the text-side directions.
Given a tangent feature $x^v\in\mathbb{R}^d$, its range coordinates and residual component are:
\begin{equation}
  a^v=x^v U_k\in\mathbb{R}^{k},\qquad
  x^v_{\mathrm{res}}=x^v - a^v U_k^\top \in\mathbb{R}^d,
  \label{eq:range_res_proj}
\end{equation}
and analogously $a^t=x^t V_k$, $x^t_{\mathrm{res}}=x^t - a^t V_k^\top$ on the text side.
The basis $(U_k,V_k)$ is computed from the \emph{real} batch and reused to project both real and synthetic features.
%, establishing a teacher--student relationship.

\smallskip
\noindent\textbf{Subspace similarity matrices.}\quad
Let $A^v, A^t$ stack the range coordinates and $X^v_{\mathrm{res}}, X^t_{\mathrm{res}}$ stack the residual components for a batch.
We form separate cross-modal similarity matrices:
\begin{equation}
  G^{\mathrm{range}} = A^v (A^t)^\top \in\mathbb{R}^{n\times n},\qquad
  G^{\mathrm{res}} = X^v_{\mathrm{res}} (X^t_{\mathrm{res}})^\top \in\mathbb{R}^{n\times n},
  \label{eq:range_res_G}
\end{equation}
where $n$ is $B$ for real or $\tilde{B}$ for synthetic data.

\smallskip
\noindent\textit{Row-wise relevance distributions.}\quad
Each similarity matrix is converted into row-wise probability distributions using a relevance temperature $\tau_r>0$.
The raw similarities in Eq.~\ref{eq:range_res_G} are divided by $\tau_r$ to form logits, and the softmax within the transport loss (below) divides by $\tau_r$ again, giving an effective temperature of $\tau_r^2$:
\begin{equation}
  P^{\mathrm{range}}=\mathrm{softmax}\!\left(\frac{G^{\mathrm{range}}_{\mathrm{real}}}{\tau_r^2}\right),\qquad
  P^{\mathrm{res}}=\mathrm{softmax}\!\left(\frac{G^{\mathrm{res}}_{\mathrm{real}}}{\tau_r^2}\right),
  \label{eq:row_softmax}
\end{equation}
with synthetic counterparts $Q^{\mathrm{range}},Q^{\mathrm{res}}$ from $G^{\mathrm{range}}_{\mathrm{syn}}$ and $G^{\mathrm{res}}_{\mathrm{syn}}$.
Real distributions $P$ are stop-gradient targets.

\smallskip
\noindent\textit{Entropy-regularized set matching.}\quad
Given a synthetic row distribution $q_i$ (the $i$-th row of $Q$) and a real row distribution $p_j$ (the $j$-th row of $P$), we measure discrepancy by KL divergence:
\vspace{-2mm}
\begin{equation}
  \mathrm{D}(q_i,p_j)=\sum_{m} q_{im} \log\frac{q_{im}}{p_{jm}}.
  \label{eq:row_kl}
\vspace{-1mm}
\end{equation}
Stacking all pairwise divergences yields a cost matrix $\Gamma\in\mathbb{R}^{\tilde{B}\times B}$ with $\Gamma_{ij}=\mathrm{D}(Q_{i,:},P_{j,:})$.
Since there is no prescribed one-to-one correspondence between synthetic and real rows, we compute a soft coupling $T\in\mathbb{R}^{\tilde{B}\times B}$ via entropy-regularization~\cite{cuturi2013sinkhorn}: % optimal transport:
\vspace{-2mm}
\begin{equation}
  T
  =
  \arg\min_{T\ge 0}
  \;\langle T,\Gamma\rangle - \varepsilon\,\mathcal{H}(T)
  \quad
  \text{s.t.}
  \quad
  T\mathbf{1}=\tfrac{1}{\tilde{B}}\mathbf{1},\;
  T^\top\mathbf{1}=\tfrac{1}{B}\mathbf{1},
  \label{eq:entropic_coupling}
\vspace{-1mm}
\end{equation}
where $\varepsilon>0$ is the regularization strength and
$\mathcal{H}(T)=-\sum_{i,j}T_{ij}\log T_{ij}$.
This is solved by Sinkhorn--Knopp iterations~\cite{cuturi2013sinkhorn} on the Gibbs kernel $K_{ij}=\exp(-\Gamma_{ij}/\varepsilon)$.
The coupling $T$ is computed under stop-gradient on $\Gamma$, so gradients flow only through the KL cost terms.
The transport-weighted cost gives a one-directional loss:
\vspace{-2mm}
\begin{equation}
  \mathcal{L}_{\rightarrow}(G_{\mathrm{real}},G_{\mathrm{syn}})
  =
  \tau_r^2 \sum_{i,j} T_{ij}\,\Gamma_{ij},
  \label{eq:set_match_loss}
\vspace{-1mm}
\end{equation}
where the $\tau_r^2$ factor compensates for the temperature scaling applied during distribution computation.
We enforce bidirectionality by averaging the image-to-text and text-to-image directions:
\vspace{-2mm}
\begin{equation}
  \mathcal{L}_{\mathrm{match}}(G_{\mathrm{real}},G_{\mathrm{syn}})
  =
  \tfrac{1}{2}\bigl[
    \mathcal{L}_{\rightarrow}(G_{\mathrm{real}},G_{\mathrm{syn}})
    +
    \mathcal{L}_{\rightarrow}(G_{\mathrm{real}}^\top,G_{\mathrm{syn}}^\top)
  \bigr].
  \label{eq:bi_match_loss}
\vspace{-1mm}
\end{equation}
Applying Eq.~\ref{eq:bi_match_loss} to range and residual similarities yields
$\mathcal{L}_{\mathrm{match}}^{\mathrm{range}}$ and $\mathcal{L}_{\mathrm{match}}^{\mathrm{res}}$.

\noindent\textbf{Range loss.}\quad
The range loss combines relevance matching with an energy regularizer that prevents the synthetic range component from collapsing.
Define the projected cross-covariance blocks
$\widehat{C}^{\mathrm{range}}_{\mathrm{real}}=U_k^\top C_{\mathrm{real}} V_k$ and
$\widehat{C}^{\mathrm{range}}_{\mathrm{syn}} =U_k^\top C_{\mathrm{syn}} V_k$, both in $\mathbb{R}^{k\times k}$,
and their mean squared energies
\begin{equation}
  e_{\mathrm{real}}=\frac{1}{k^2}\|\widehat{C}^{\mathrm{range}}_{\mathrm{real}}\|_F^2,\qquad
  e_{\mathrm{syn}} =\frac{1}{k^2}\|\widehat{C}^{\mathrm{range}}_{\mathrm{syn}}\|_F^2.
\end{equation}
The regularizer penalizes synthetic energy only when it falls below the real energy:
\vspace{-2mm}
\begin{equation}
  \mathcal{L}_{\mathrm{reg}}^{\mathrm{range}}
  =
  \frac{\max(0,\; e_{\mathrm{real}}-e_{\mathrm{syn}})}{e_{\mathrm{real}}+\epsilon},
  \label{eq:range_expand}
\end{equation}
where $\epsilon>0$ is a small constant for numerical stability.
This one-sided penalty ensures the synthetic data maintains at least as much coupling energy as the real data in the top-$k$ subspace, without penalizing cases where synthetic energy exceeds the real.
The grouped range loss is then:
\vspace{-1.5mm}
\begin{equation}
  \mathcal{L}_{\mathrm{range}}
  =
  \mathcal{L}_{\mathrm{match}}^{\mathrm{range}}
  +
  \mathcal{L}_{\mathrm{reg}}^{\mathrm{range}}.
  \label{eq:range_grouped}
\end{equation}

\noindent\textbf{Residual loss.}\quad
The residual subspace captures cross-modal interactions outside the top-$k$ range.
Matching these interactions can sharpen ranking margins, but if residual energy dominates range energy, the distilled data over-represents weakly correlated directions at the expense of the dominant structure.
The residual loss thus pairs a matching term with a compression regularizer.
We form the residual cross-covariance of synthetic data by projecting out the range on both sides:
\vspace{-1.5mm}
\begin{equation}
  C^{\mathrm{res}}_{\mathrm{syn}}
  =
  (I - U_k U_k^\top)\, C_{\mathrm{syn}}\, (I - V_k V_k^\top) \;\in\mathbb{R}^{d\times d},
  \label{eq:residual_xcov}
\end{equation}

\noindent where $I$ is the $d\times d$ identity.
We let the residual energy defined as $e_{\mathrm{res}}=\frac{1}{d^2}\|C^{\mathrm{res}}_{\mathrm{syn}}\|_F^2$ and the ratio $r=e_{\mathrm{res}} / (e_{\mathrm{syn}}+\epsilon)$, measuring how much synthetic coupling energy lies outside the range. The compression regularizer is defined as:
\vspace{-1.5mm}
\begin{equation}
  \mathcal{L}_{\mathrm{reg}}^{\mathrm{residual}}
  =
  r + \max(0,\; r-1).
  \label{eq:residual_comp}
\end{equation}
The first term provides a steady gradient shrinking $r$ toward zero, while the second activates an additional penalty when $r>1$ (residual energy exceeds range energy).
We then define the grouped residual loss in Eq.~\ref{eq:res_grouped} where $\lambda_{\mathrm{comp}}$ controls the relative strength of compression versus matching within the residual subspace.
\vspace{-3mm}
\begin{equation}
  \mathcal{L}_{\mathrm{residual}}
  =
  \mathcal{L}_{\mathrm{match}}^{\mathrm{residual}}
  +
  \lambda_{\mathrm{comp}}\,\mathcal{L}_{\mathrm{reg}}^{\mathrm{residual}}.
  \label{eq:res_grouped}
\end{equation}

\vspace{-3mm}

\subsection{Final Distillation Training Objective}
\label{sec:final_objective}

The total distillation objective is then defined as:
\vspace{-2mm}
\begin{equation}
  \mathcal{L}_{\mathrm{total}}
  =
  \mathcal{L}_{\mathrm{hITC}}
  \;+\;
  \lambda_{\mathrm{range}}\,\mathcal{L}_{\mathrm{range}}
  \;+\;
  \lambda_{\mathrm{residual}}\,\mathcal{L}_{\mathrm{residual}}.
  \label{eq:total_loss}
\end{equation}

\noindent Here, $\lambda_{\mathrm{range}}$ and $\lambda_{\mathrm{residual}}$ weight the range and residual subspace losses relative to the contrastive term.
Expanding Eq.~\ref{eq:range_grouped} and Eq.~\ref{eq:res_grouped}, the three scalar hyperparameters $(\lambda_{\mathrm{range}}, \lambda_{\mathrm{residual}}, \lambda_{\mathrm{comp}})$ control four loss components: range matching and its regularizer share $\lambda_{\mathrm{range}}$, residual matching is scaled by $\lambda_{\mathrm{res}}$, and residual compression by $\lambda_{\mathrm{residual}}\lambda_{\mathrm{comp}}$.
By default, $\lambda_{\mathrm{range}}$=$0.8$, $\lambda_{\mathrm{residual}}$=$0.4$, $\lambda_{\mathrm{comp}}$=$0.1$.

\smallskip
\noindent\textbf{Training procedure.}\quad
Only the synthetic parameters $(\tilde{x}_j, \tilde{\mathbf{E}}_j)$ are updated by gradient descent by Eq.~\ref{eq:total_loss}.
The encoder weights $(\theta_v,\theta_t,\omega_v,\omega_t)$ are initialized from pretrained checkpoints and held fixed during the synthetic update step.
Following the online distillation protocol of~\cite{lee2025covmatch}, each distillation iteration consists of multiple outer-loop steps.
Within each outer step: (1) a real batch is sampled and its features are computed with stop-gradient, (2) a synthetic sub-batch of size $\tilde{B}\le M$ is sampled
% (with uniform random subsampling when $M>\tilde{B}$)
, (3) distillation loss is computed and backpropagated through the synthetic pathway only, and (4) the synthetic parameters are updated by SGD.
For each outer-loop step, the network is updated for one inner-loop step on real data, providing a slowly evolving latent feature landscape for enhanced generalization in the next outer step.
Alg.~\ref{alg:raha} summarizes the procedure.

\begin{algorithm}[t]
\caption{Distilling data with Rank-Aware Hyperbolic Alignment}
\label{alg:raha}
\centering
\resizebox{0.8\linewidth}{!}{%
\begin{minipage}{\linewidth}
\begin{algorithmic}[1]
  \REQUIRE Real dataset $\mathcal{D}_{\mathrm{real}}$, synthetic budget $M$, curvature $c$, scale $s$, \\
  \quad\quad\quad\; outer steps $N_{\mathrm{out}}$, inner steps $N_{\mathrm{in}}$, iterations $T$, batch sizes $B,\tilde{B}$
  \STATE Initialize $\mathcal{D}_{\mathrm{syn}}=\{(\tilde{x}_j,\tilde{\mathbf{E}}_j,\tilde{\mathbf{m}}_j)\}_{j=1}^{M}$ from real samples
  \STATE Initialize encoder $\Psi$ from pretrained weights
  \FOR{$t=1$ \TO $T$}
    % \STATE Initialize encoder $\Psi$ to pretrained weights to avoid drift and stabilize rank selection
    \FOR{$\ell=1$ \TO $N_{\mathrm{out}}$}
      \STATE \textbf{Synthetic update:}
      \STATE \quad Sample real batch $\mathcal{B}$ and synthetic sub-batch $\tilde{\mathcal{B}}$ 
      \STATE \quad Compute projected features $(z^v,z^t)$ for both batches via $\Psi$\hfill\COMMENT{Eq.~\ref{eq:z_def}}
      \STATE \quad Lift synthetic features to hyperboloid via Exp, compute $\mathcal{L}_{\mathrm{hITC}}$\hfill\COMMENT{Eqs.~\ref{eq:lift}--\ref{eq:hITC}}
      \STATE \quad Re-map all features to tangent space via Log\hfill\COMMENT{Eq.~\ref{eq:log_map}}
      \STATE \quad Compute $C_{\mathrm{real}}, C_{\mathrm{syn}}$, SVD of $C_{\mathrm{real}}$, and rank $k$\hfill\COMMENT{Eqs.~\ref{eq:xcov_real}--\ref{eq:rank_energy}}
      \STATE \quad Project into range and residual, form $G^{\mathrm{range}},G^{\mathrm{res}}$\hfill\COMMENT{Eqs.~\ref{eq:range_res_proj}--\ref{eq:range_res_G}}
      \STATE \quad Compute relevance distributions and Sinkhorn coupling\hfill\COMMENT{Eqs.~\ref{eq:row_softmax}--\ref{eq:entropic_coupling}}
      \STATE \quad Compute $\mathcal{L}_{\mathrm{range}}$ and $\mathcal{L}_{\mathrm{res}}$\hfill\COMMENT{Eqs.~\ref{eq:range_grouped},~\ref{eq:res_grouped}}
      \STATE \quad Update $\mathcal{D}_{\mathrm{syn}}$ by SGD on $\mathcal{L}_{\mathrm{total}}$\hfill\COMMENT{Eq.~\ref{eq:total_loss}}
      \STATE \textbf{Model update:} Train $\Psi$ for $N_{\mathrm{in}}$ steps on real data 
      % \IF{$\ell < N_{\mathrm{out}}$}
      % \ENDIF
    \ENDFOR
  \ENDFOR
  \RETURN $\mathcal{D}_{\mathrm{syn}}^{\ast}$
\end{algorithmic}
\end{minipage}%
}
\end{algorithm}
\vspace{-3mm}

\section{Experiments}
\label{sec:experiments}

\subsection{Experimental Setup}
\label{subsec:exp_setup}

\paragraph{Datasets and splits.}
We evaluate multimodal dataset distillation on three canonical image--caption benchmarks for
bidirectional retrieval: Flickr8k~\cite{hodosh2013flickr8k}, Flickr30k~\cite{young2014flickr}, and
MS COCO~\cite{COCO}. We use the standard retrieval splits popularized by Karpathy \emph{et al.}~\cite{karpathy2015deep}.
Unless otherwise noted, COCO results are reported on the 5k-image test split (rather than the 1k subset),
which is the most common setting in recent MDD evaluations.
Dataset statistics are summarized in Table~\ref{tab:dataset_stats}.

\begin{table}[!htbp]
    \centering
    \caption{\textbf{Dataset statistics for image--text retrieval.}
    Splits follow the standard retrieval protocol~\cite{karpathy2015deep}.
    Each image is annotated with five human-written captions.}
    \vspace{-3mm}
    \label{tab:dataset_stats}
    \scriptsize    
    \renewcommand{\arraystretch}{1.15}
    \renewcommand{\aboverulesep}{3pt}
    \renewcommand{\belowrulesep}{3pt}
    \setlength{\tabcolsep}{4pt}
    \resizebox{.9\linewidth}{!}{
    \begin{tabular}{l c c c p{0.40\linewidth}}
        \toprule
        Dataset & Year & \#Images (Train/Val/Test) & Caps/img & Caption characteristics \\
        \midrule
        Flickr8k~\cite{hodosh2013flickr8k}  & 2013 & 6k / 1k / 1k & 5 &
        Short, single-sentence human captions describing everyday scenes; relatively clean and small-scale. \\
        Flickr30k~\cite{young2014flickr} & 2014 & 29k / 1k / 1k & 5 &
        Crowd-sourced, concise descriptions with broader visual diversity (people, objects, actions, relations). \\
        MS COCO~\cite{COCO}   & 2014 & 113k / 5k / 5k & 5 &
        Large-scale, object-centric captions with richer compositionality and vocabulary; the most diverse of the three. \\
        \bottomrule
    \end{tabular}}
    \vspace{-6mm}
\end{table}

\paragraph{Task and metrics.}\quad
We use bidirectional image--text retrieval as the main probe of cross-modal alignment.
We report Recall@K for $K\in\{1,5,10\}$ in both directions:
text-to-image (T$\rightarrow$I, denoted \textbf{IR@K}) and image-to-text (I$\rightarrow$T, denoted \textbf{TR@K}).
Retrieval similarities are computed by \textbf{dot product} in the shared embedding space.
For fair comparison, we evaluate \emph{all} methods using the same similarity function.

\paragraph{Synthetic budget.}\quad
We distill a compact synthetic set of $N$ image--text queries with $N \in \{100,\, 200,\, 500\}$. Each query consists of (i) one image of size $3\times224\times224$ optimized in pixel space and
(ii) one \emph{continuous} 768-dimensional text embedding optimized directly in embedding space
(i.e., we synthesize embeddings rather than discrete tokens), following the standard embedding-level
MDD setup~\cite{wu2024vldistill,xu2024lors}.
These budgets correspond to strong compression regimes (e.g., $N{=}100$ is below 1\% of COCO training images).

\paragraph{Baselines and network architecture.}\quad
We compare RAHA to (i) coreset selection under the same budget (Random, Herding, K-Center, Forgetting)~\cite{c-herd,c-kcenter,c-forget}, and (ii) vision--language distillation methods, including trajectory-matching approaches (\cite{wu2024vldistill}, \cite{cui2023tesla}, \cite{xu2024lors}, \cite{zhang2025repblend}) and distribution/statistics matching \cite{lee2025covmatch}. All methods use the same budget, architecture, and evaluation protocol.

We use the network composed from NFNet~\cite{brock2021nfnet} and BERT~\cite{devlin2019bert}, each followed by a lightweight linear projection head to a shared $d$-dimensional embedding space~\cite{wu2024vldistill, xu2024lors, zhang2025repblend, lee2025covmatch}. 

\paragraph{Distillation protocol.}
We optimize the synthetic images and synthetic text embeddings while using an \emph{online} surrogate model to compute distillation gradients.
Each outer distillation iteration alternates between two stages: (i) an offline synthetic update stage, where we update $\mathcal{D}_{\mathrm{syn}}$ for $N_{\mathrm{out}}$ (\eg, 50) gradient steps using the current model state, and (ii) online model update stage for $N_{\mathrm{in}}$ (\eg, 1), where we update the trainable retrieval model for one step on real data.
This alternating schedule intentionally varies the surrogate model state during distillation, exposing the synthetic set to a shifting encoder geometry and improving the generality of the distilled pairs across latent-space configurations.
RAHA computes its relevance matching terms in the hyperbolic representation induced by Lorentz lifting and tangent-space operations, and we allow a structured radial modality gap, consistent with the observation that text concepts are often more generic than images \cite{desai2023meru}.
After distillation, we train a retrieval model from scratch on the distilled set for 100 epochs and evaluate on the real test split.

\subsection{Main Results}
\label{subsec:exp_results}

\subsubsection{Image--Text Retrieval.}
%%%%%%%%%%%%%%%%%%%%%%%%%%%%%%%%%%%%%%%%%%%%%%%%%%%
\begin{table*}[t]
\centering

\caption{\textbf{Image-text retrieval} for 100, 200, and 500 synthetic pairs using the coreset methods (\textit{left}) and representative distillation methods (\textit{right}). The compression rate for \{Flickr8k, Flickr30k, and COCO\} datasets are approximately \{1.7\%, 0.3\%, 0.8\textperthousand\}, \{3.3\%, 0.7\%, 1.7\textperthousand\}, \{8.3\%, 1.7\%, 4.4\textperthousand\} for 100, 200, and 500 pairs, respectively. 
% Best and runner-up results are indicated in \textbf{boldface} and \underline{underline}, respectively.
}
    \vspace{-3mm}
    \label{tab:retrieval}
    \renewcommand{\arraystretch}{1.15}
    \renewcommand{\aboverulesep}{12pt}
    \renewcommand{\belowrulesep}{12pt}
    \setlength{\tabcolsep}{4pt}
    \scriptsize
    \fontsize{28}{28}\selectfont
    % \fontsize{14}{14}\selectfont
    \resizebox{.99\linewidth}{!}{
    \begin{tabular}{cc|cccccccccccc|ccccccccccccccccccccc}
    \toprule
    
    &&
    \multicolumn{12}{c}{\textbf{Coreset sampling}} &
    \multicolumn{18}{c}{\textbf{Distillation methods}} \\
    \cmidrule(lr){3-14}\cmidrule(lr){15-32}
    \cmidrule(lr){15-23}\cmidrule(lr){24-29}
    
    &&
    &&&&&&&&&&&&
    \multicolumn{9}{c}{Matching Training Trajectories (MTT)} 
    &
    \multicolumn{6}{c}{Distribution Matching (DM)} \\
    % \cmidrule(lr){15-26}\cmidrule(lr){27-32}
    \cmidrule(lr){15-23}\cmidrule(lr){24-29}
    
    & &
    \multicolumn{3}{c}{Random} &
    \multicolumn{3}{c}{Herding~\cite{c-herd}} &
    \multicolumn{3}{c}{K-Center~\cite{c-kcenter}} &
    \multicolumn{3}{c|}{Forgetting~\cite{c-forget}} &
    
    \multicolumn{3}{c}{MTT-VL~\cite{wu2024vldistill}} &
    \multicolumn{3}{c}{TESLA$_{\textrm{WBCE}}$~\cite{cui2023tesla}} & % % $_{\textrm{WBCE}}$
    \multicolumn{3}{c}{LoRS~\cite{xu2024lors}} &
    % \multicolumn{3}{c}{RepBlend~\cite{zhang2025repblend}} &
    \multicolumn{3}{c}{CovMatch~\cite{lee2025covmatch}} &
    \multicolumn{3}{c}{\cellcolor{eccvblue!20}Ours}\\
    
    \cmidrule(lr){3-5}\cmidrule(lr){6-8}\cmidrule(lr){9-11}\cmidrule(lr){12-14}
    \cmidrule(lr){15-17}\cmidrule(lr){18-20}\cmidrule(lr){21-23}\cmidrule(lr){24-26}\cmidrule(lr){27-29}
    % \cmidrule{30-32}
    
    \multirow[b]{-2}{*}{\rotatebox{90}{Data}} &
    \multirow[b]{-2}{*}{\rotatebox{90}{\# Pairs}} &
    IR & TR & \textbf{Mean} &
    IR & TR & \textbf{Mean} &
    IR & TR & \textbf{Mean} &
    IR & TR & \textbf{Mean} &
    
    IR & TR & \textbf{Mean} &
    IR & TR & \textbf{Mean} &
    IR & TR & \textbf{Mean} &
    % IR & TR & \textbf{Mean} &
    IR & TR & \textbf{Mean} &
    IR & TR & \textbf{Mean} \\
    \midrule
    
    \multirow{3}{*}{\rotatebox{90}{Flickr8k}} &
    100  &
    3.5 & 7.8 & \cellcolor{eccvblue!20}5.7 &   % Random
    4.7 & 8.3 & \cellcolor{eccvblue!20}6.5 &   % Herding    
    5.0 & 8.6 & \cellcolor{eccvblue!20}6.8 &   % K-Center
    4.1 & 4.9 & \cellcolor{eccvblue!20}5.1 &   % Forgetting

    3.9 & 6.2 & \cellcolor{eccvblue!20}5.1 &       % MTT-VL
    4.6 & 14.1 & \cellcolor{eccvblue!20}9.4 &      % TESLA WBCE
    17.3 & 21.5 & \cellcolor{eccvblue!20}19.4 &    % LoRS
    % 21.9 & 30.6 & \cellcolor{eccvblue!20}26.3 &    % RepBlend)
    18.4 & 22.4 & \cellcolor{eccvblue!20}20.4 &    % CovMatch
    19.0 & 21.9 & \cellcolor{eccvblue!20}20.4 \\   % Ours
    
    & 200  &
    7.8 & 11.8 & \cellcolor{eccvblue!20}9.8 & 
    5.5 & 11.9 & \cellcolor{eccvblue!20}\cellcolor{eccvblue!20}8.7 & 
    8.8 & 12.8 & \cellcolor{eccvblue!20}10.8 & 
    6.6 & 9.4 & \cellcolor{eccvblue!20}8.0 & 
    
    7.0 & 10.1 & \cellcolor{eccvblue!20}8.6 & 
    4.8 & 18.5 & \cellcolor{eccvblue!20}11.7 & 
    19.5 & 24.7 & \cellcolor{eccvblue!20}22.1 & 
    % 26.2 & 34.7 & \cellcolor{eccvblue!20}30.4 & 
    17.4 & 20.3 & \cellcolor{eccvblue!20}18.8 & % covmatch    
    22.7 & 27.9 & \cellcolor{eccvblue!20}25.3 \\
    
    & 500  &
    12.6 & 18.1 & \cellcolor{eccvblue!20}15.4 & 
    12.0 & 16.3 & \cellcolor{eccvblue!20}14.1 & 
    12.8 & 17.9 & \cellcolor{eccvblue!20}15.4 & 
    15.1 & 19.7 & \cellcolor{eccvblue!20}17.4 & 
    
    12.7 & 17.3 & \cellcolor{eccvblue!20}15.0 & 
    8.5 & 18.5 & \cellcolor{eccvblue!20}13.5 &
    20.7 & 29.2 & \cellcolor{eccvblue!20}25.0 & 
    % 18.4 & 33.3 & \cellcolor{eccvblue!20}25.9 & 
    23.8 & 28.0 & \cellcolor{eccvblue!20}25.9 & % covmatch
    27.7 & 33.6 & \cellcolor{eccvblue!20}30.7 \\
    
    % & 1000 &
    % -- & -- & -- &
    % -- & -- & -- &
    % -- & -- & -- &
    % -- & -- & -- &
    % -- & -- & -- &
    % -- & -- & -- &
    % -- & -- & -- &
    % -- & -- & -- &
    % -- & -- & -- \\
    
    \midrule
    
    \multirow{3}{*}{\rotatebox{90}{Flickr30k}} &
    100  &
    7.4 & 9.9 & \cellcolor{eccvblue!20}8.6 &   % random
    7.9 & 9.5 & \cellcolor{eccvblue!20}8.7 &   % herding
    7.5 & 9.6 & \cellcolor{eccvblue!20}8.6 &   % kcenter
    2.9 & 5.0 & \cellcolor{eccvblue!20}4.0 &   % forgetting
    
    15.0 & 25.8 & \cellcolor{eccvblue!20}20.4 & 
    2.5 & 18.0 & \cellcolor{eccvblue!20}10.2 & 
    22.5 & 32.3 & \cellcolor{eccvblue!20}27.4 & 
    % 29.3 & 37.8 & \cellcolor{eccvblue!20}33.6 & 
    % 26.5 & 34.5 & \cellcolor{eccvblue!20}30.5 & % covmatch
    20.9 & 24.6 & \cellcolor{eccvblue!20}22.8 & % covmatch*    
    18.7 & 22.7 & \cellcolor{eccvblue!20}20.7 \\        % Ours 
    
    & 200  &
    11.1 & 15.1 & \cellcolor{eccvblue!20}13.1 & 
    10.9 & 14.3 & \cellcolor{eccvblue!20}12.6 & 
    10.7 & 14.6 & \cellcolor{eccvblue!20}12.6 & 
    4.2 & 6.7 & \cellcolor{eccvblue!20}5.5 & 
    
    15.4 & 26.9 & \cellcolor{eccvblue!20}21.2 & 
    1.3 & 10.2 & \cellcolor{eccvblue!20}5.8 & 
    23.5 & 35.5 & \cellcolor{eccvblue!20}29.5 & 
    % 31.7 & 41.5 & \cellcolor{eccvblue!20}36.6 & 
    % 30.6 & 38.3 & \cellcolor{eccvblue!20}34.4 & % covmatch
    20.2 & 23.7 & \cellcolor{eccvblue!20}22.0 & % covmatch*
    23.5 & 27.9 & \cellcolor{eccvblue!20}25.7 \\
    
    & 500  &
    19.7 & 25.6 & \cellcolor{eccvblue!20}22.6 & 
    19.5 & 24.2 & \cellcolor{eccvblue!20}21.8 & 
    20.4 & 26.9 & \cellcolor{eccvblue!20}23.7 & 
    8.9 & 11.7 & \cellcolor{eccvblue!20}10.3 & 
    
    18.9 & 31.0 & \cellcolor{eccvblue!20}25.0 & 
    7.0 & 14.7 & \cellcolor{eccvblue!20}10.9 & 
    26.8 & 36.3 & \cellcolor{eccvblue!20}31.6 & 
    % 28.5 & 47.5 & \cellcolor{eccvblue!20}43.0 & 
    % 34.8 & 42.0 & \cellcolor{eccvblue!20}38.4 & % covmatch
    26.3 & 31.5 & \cellcolor{eccvblue!20} 28.9 &  % covmatch*
    30.0 & 35.9 & \cellcolor{eccvblue!20}32.9 \\
    
    \midrule
    
    \multirow{3}{*}{\rotatebox{90}{COCO}} &
    100  &
    2.9 & 4.0 & \cellcolor{eccvblue!20}3.4 & 
    3.0 & 4.0 & \cellcolor{eccvblue!20}3.5 & 
    3.3 & 4.2 & \cellcolor{eccvblue!20}3.8 & 
    1.4 & 2.7 & \cellcolor{eccvblue!20}2.1 & 
    
    5.4 & 9.4 & \cellcolor{eccvblue!20}7.4 & 
    1.0 & 7.7 & \cellcolor{eccvblue!20}4.4 & 
    7.0 & 11.7 & \cellcolor{eccvblue!20}9.4 & 
    % 13.4 & 17.0 & \cellcolor{eccvblue!20}15.2 & 
    % 10.3 & 12.7 & \cellcolor{eccvblue!20}11.5 & % covmatch
    6.5 & 7.4 & \cellcolor{eccvblue!20} 7.0 &  % covmatch*
    6.6 & 7.7 & \cellcolor{eccvblue!20}7.2 \\
    
    & 200  &
    4.7 & 6.4 & \cellcolor{eccvblue!20}5.6 & 
    4.8 & 6.5 & \cellcolor{eccvblue!20}5.6 & 
    5.1 & 6.7 & \cellcolor{eccvblue!20}5.9 & 
    2.8 & 3.9 & \cellcolor{eccvblue!20}3.3 & 
    
    6.8 & 11.5 & \cellcolor{eccvblue!20}9.2 & 
    0.3 & 3.0 & \cellcolor{eccvblue!20}1.7 & 
    9.1 & 13.7 & \cellcolor{eccvblue!20}11.4 & 
    % 18.4 & 20.3 & \cellcolor{eccvblue!20}19.4 &  % RepBlend
    % 13.0 & 16.5 & \cellcolor{eccvblue!20}14.8 & % covmatch
    7.4 & 9.2 & \cellcolor{eccvblue!20} 8.3 &  % covmatch*
    9.3 & 11.0 & \cellcolor{eccvblue!20}10.2\\
    
    & 500  &
    8.6 & 11.5 & \cellcolor{eccvblue!20}10.1 & 
    8.6 & 11.0 & \cellcolor{eccvblue!20}9.8 & 
    8.9 & 12.0 & \cellcolor{eccvblue!20}10.5 & 
    4.9 & 7.8 & \cellcolor{eccvblue!20}6.4 & 
    
    9.1 & 16.1 & \cellcolor{eccvblue!20}12.6 & 
    3.7 & 5.9 & \cellcolor{eccvblue!20}4.8 & 
    9.7 & 17.2 & \cellcolor{eccvblue!20}13.5 & 
    % 18.9 & 20.6 & \cellcolor{eccvblue!20}19.8 & 
    % 17.2 & 22.1 & \cellcolor{eccvblue!20}19.6 &  % covmatch
    9.9 & 8.3 & \cellcolor{eccvblue!20} 11.2 &  % covmatch*
    12.6 & 14.9 & \cellcolor{eccvblue!20}13.7 \\
    
    \bottomrule
    \end{tabular}
    }  
\vspace{-5mm}
\end{table*}

We report bidirectional retrieval performance on Flickr8k, Flickr30k, and COCO for budgets $N\in\{100,200,500\}$ in Table~\ref{tab:retrieval}.
Across all datasets and budgets, RAHA consistently outperforms coreset selection under the same budget, highlighting the benefit of optimizing synthetic pairs rather than selecting real pairs.
Compared to trajectory-matching baselines (MTT-VL, TESLA, LoRS, RepBlend), RAHA is competitive across budgets while operating in a distribution-matching regime that does not require storing expert trajectories.
On Flickr8k, RAHA scales strongly with $N$ and attains the best overall mean recall at $N{=}500$, indicating that subspace-conditioned relevance matching preserves retrieval-relevant ranking structure even under extreme compression.
On Flickr30k and COCO, RAHA remains comparable to strong trajectory-based baselines and improves with increasing $N$, suggesting that relevance distillation benefits from additional synthetic capacity on more diverse datasets. See comparison with EDGE~\cite{zhao2025edge} in Appendix.

\begin{figure}[!t]
    \vspace{1mm}
    \centering
    \includegraphics[width=\linewidth]{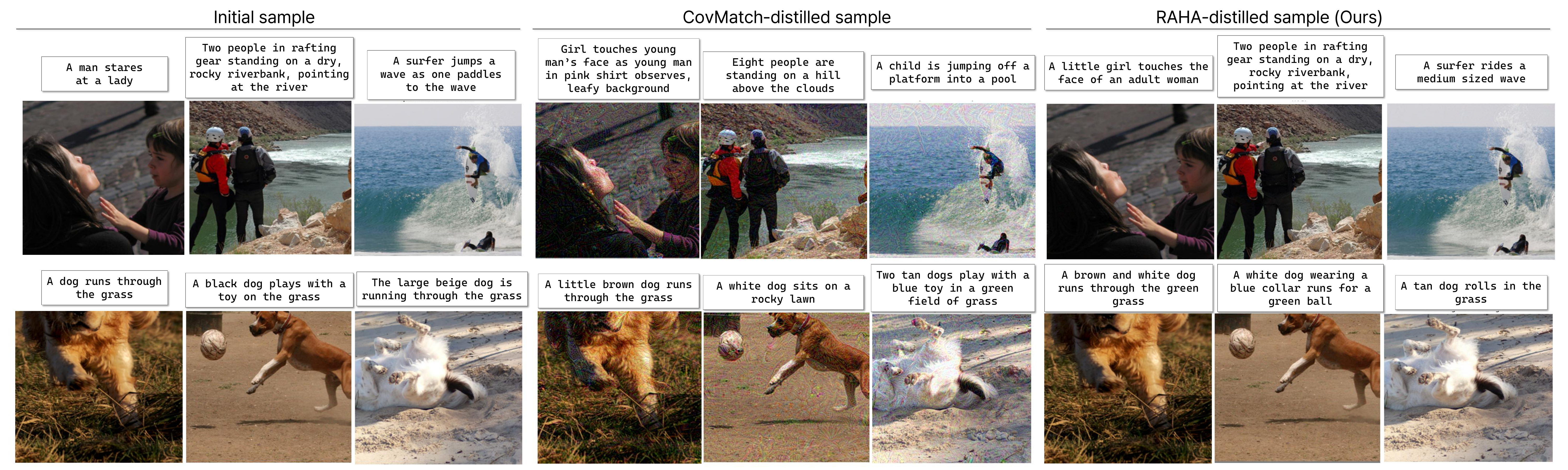}
    \vspace{-7mm}
    \caption{\textbf{Qualitative synthesized pairs.} Representative samples at initialization (\emph{left}), after CovMatch (\emph{middle}),
    and after RAHA distillation (\emph{right}). Please zoom in for details and view in color.}
    \label{fig:qualitative}
    \vspace{-6mm}
\end{figure}

Relative to the strongest distribution/statistics matching baseline, CovMatch~\cite{lee2025covmatch}, RAHA trades some absolute retrieval performance in the smallest-budget regime for a different inductive bias.
CovMatch directly aligns Euclidean cross-covariances and feature-level statistics, which is highly effective when $N$ is extremely small.
RAHA instead matches \emph{row-wise relevance distributions} after extracting a rank-adaptive range/residual decomposition from real batch structure, explicitly controlling how alignment capacity is allocated across coupled and weakly-coupled directions.
This distinction is reflected most clearly in cross-architecture transfer and robustness (Table~\ref{tab:robustness}), and is further supported by qualitative differences in the synthesized pairs (Fig.~\ref{fig:qualitative}).

\paragraph{Qualitative comparison.}
Fig.~\ref{fig:qualitative} contrasts representative synthesized pairs.
CovMatch outputs often retain visually salient \emph{residual artifacts} such as high-frequency speckling and banding, even when coarse layout is plausible.
Such artifacts are consistent with satisfying global second-order alignment through structured noise that perturbs features without improving perceptual realism.
In contrast, RAHA yields cleaner textures and more natural edges, suggesting that range--residual relevance matching suppresses spurious degrees of freedom that manifest as visible artifacts.
RAHA also better preserves fine-grained image--text semantics in several cases, where CovMatch drifts to mismatched captions (e.g., incorrect scenes/actions), while RAHA retains the correct relational content and discriminative attributes.
Overall, the qualitative evidence aligns with RAHA's design goal: preserve dominant coupled structure while explicitly controlling weakly-coupled residual interactions.

\subsection{Architectural Transfer and Robustness}

\begin{table}[!t]
\centering
\caption{Comparison on Flickr8k~\cite{hodosh2013flickr8k}: (i) \textbf{Cross-architecture generalization} averaged over IR/TR@K=\{1,5,10\}. Source-model results marked with `$^{\ast}$' are not averaged; best average results are in \textbf{boldface}, and runner-up in \underline{underline}. (a)--(d): NFNet, NF-ResNet, NF-RegNet, ViT-B. (ii) \textbf{Robustness to noise ($\delta$) by modality}: (a)--(c): JPEG~\cite{JPEG} (75\%)/Bit Quantization~\cite{xu2017bitreduction} (4-bit), Additive White Gaussian Noise (AWGN) ($\sigma=0.01$), 10-PGD~\cite{madry2017pgd} (step size=2).}
\vspace{-3mm}
\label{tab:robustness}

\renewcommand{\arraystretch}{1.15}
\renewcommand{\aboverulesep}{2pt}
\renewcommand{\belowrulesep}{2pt}
\setlength{\tabcolsep}{5pt}
\scriptsize
\fontsize{12}{12}\selectfont

\resizebox{.99\linewidth}{!}{%
% 2 (left) + 9 (panel a, now includes ViT) + 11 (panel b)  = 22 columns total
\begin{tabular}{c r c c c c c c c c c | c c c c c c c c c c c}
\toprule
\multirow{3}{*}{\rotatebox[origin=c]{90}{\# Pairs}} 
&
&
\multicolumn{9}{c}{\bfseries (i) Cross-architecture generalization} 
&
\multicolumn{10}{c}{\bfseries (ii) Robustness to noise} \\
\cmidrule(lr){3-11}\cmidrule(lr){12-20}
&
\multirow{1}{*}{Text $\varepsilon$}
&
% cross-arch
\multicolumn{4}{c}{BERT~\cite{devlin2019bert}} &
\multicolumn{4}{c}{DistilBERT~\cite{sanh2019distilbert}} &
&
% robustness
\multicolumn{3}{c}{Image-side $\delta$} & 
&
\multicolumn{3}{c}{Text-side $\delta$} &  
\\
\cmidrule(lr){3-6}\cmidrule(lr){7-10}\cmidrule(lr){12-14}\cmidrule(lr){16-18}

&
\multirow{1}{*}{Image $\varepsilon$}
&
(a) & (b) & (c) & (d) 
&
(a) & (b) & (c) & (d) & \cellcolor{eccvblue!20}Mean
&
(a) & (b) & (c) & \cellcolor{eccvblue!20}Mean
&
(a) & (b) & (c) &  \cellcolor{eccvblue!20}Mean
\\
\midrule

% ===================== 100 =====================
\multirow{2}{*}{\rotatebox[origin=c]{90}{100}}

& CovMatch~\cite{lee2025covmatch}
& 20.4$^{\ast}$ & 6.4 & 6.3 & 7.7
& 9.6 & 6.8 & 6.3 & 7.8
& \cellcolor{eccvblue!20}7.3
& 6.0 & 6.0 & 1.9 & \cellcolor{eccvblue!20}4.6
& 3.8 & 10.1 & 0.0
& \cellcolor{eccvblue!20}4.6 \\

& \cellcolor{eccvblue!20}Ours
& \cellcolor{eccvblue!20}20.4$^{\ast}$ 
& \cellcolor{eccvblue!20}5.1& \cellcolor{eccvblue!20}5.4 & \cellcolor{eccvblue!20}6.3
& \cellcolor{eccvblue!20}9.9 & \cellcolor{eccvblue!20}6.7 & \cellcolor{eccvblue!20}6.7 & \cellcolor{eccvblue!20}8.0
& \cellcolor{eccvblue!20}6.9
 & \cellcolor{eccvblue!20}4.9 & \cellcolor{eccvblue!20}5.6 & \cellcolor{eccvblue!20}1.8 & \cellcolor{eccvblue!20}4.1
 & \cellcolor{eccvblue!20}3.1 & \cellcolor{eccvblue!20}8.1 & \cellcolor{eccvblue!20}0.0 & \cellcolor{eccvblue!20}3.7\\
\midrule

% ===================== 200 =====================
\multirow{2}{*}{\rotatebox[origin=c]{90}{200}}
& CovMatch~\cite{lee2025covmatch}
& 18.8$^{\ast}$ & 7.2 & 6.3 & 6.7
& 9.3 & 7.2 & 6.7 & 7.2
& \cellcolor{eccvblue!20}7.2
& 5.0 & 5.7 & 1.9  & \cellcolor{eccvblue!20}4.2 
 & 3.3  & 8.7  & 0.0
  & \cellcolor{eccvblue!20}4.0\\

& \cellcolor{eccvblue!20}Ours
& \cellcolor{eccvblue!20}25.3$^{\ast}$ & \cellcolor{eccvblue!20}7.4 & \cellcolor{eccvblue!20}7.2 & \cellcolor{eccvblue!20}9.1
& \cellcolor{eccvblue!20}10.5 & \cellcolor{eccvblue!20}8.0& \cellcolor{eccvblue!20}7.9 & \cellcolor{eccvblue!20}10.8
& \cellcolor{eccvblue!20}8.7  
& \cellcolor{eccvblue!20}5.8 & \cellcolor{eccvblue!20}6.8 & \cellcolor{eccvblue!20}2.4
& \cellcolor{eccvblue!20}5.0 
& \cellcolor{eccvblue!20}3.6 & \cellcolor{eccvblue!20}9.8 & \cellcolor{eccvblue!20}0.0
& \cellcolor{eccvblue!20}4.5 \\
\midrule

% ===================== 500 =====================
\multirow{2}{*}{\rotatebox[origin=c]{90}{500}}
& CovMatch~\cite{lee2025covmatch}
& 25.9$^{\ast}$ & 8.1 & 7.3 & 8.9
& 12.3 & 8.4 & 7.3 & 9.1
& \cellcolor{eccvblue!20}8.7
& 8.0 & 8.3 & 3.0 & \cellcolor{eccvblue!20}6.4
& 3.9 & 11.5 & 0.0 & \cellcolor{eccvblue!20}5.1  \\

& \cellcolor{eccvblue!20}Ours
& \cellcolor{eccvblue!20}30.7$^{\ast}$ & \cellcolor{eccvblue!20}5.4 & \cellcolor{eccvblue!20}11.3 & \cellcolor{eccvblue!20}14.6
& \cellcolor{eccvblue!20}15.4 & \cellcolor{eccvblue!20}12.5 & \cellcolor{eccvblue!20}12.8 & \cellcolor{eccvblue!20}16.9
& \cellcolor{eccvblue!20}12.7
& \cellcolor{eccvblue!20}8.0 & \cellcolor{eccvblue!20}9.5 & \cellcolor{eccvblue!20}3.0 & \cellcolor{eccvblue!20}6.8
& \cellcolor{eccvblue!20}4.4 & \cellcolor{eccvblue!20}15.0 & \cellcolor{eccvblue!20}0.0 & \cellcolor{eccvblue!20}6.4 \\

\bottomrule
\end{tabular}
}
\vspace{-3mm}
\end{table}

To test whether distilled pairs capture transferable cross-modal relevance rather than overfitting the source encoders used during distillation, we follow a cross-architecture protocol.
We distill with a fixed source configuration and then retrain/evaluate retrieval models that replace either the image backbone or the text encoder while keeping the distilled set unchanged.
We also evaluate robustness via the degradation metric $\delta$ under common perturbations applied to either modality.

\paragraph{Across budgets.}
At $N{=}100$, RAHA is competitive with CovMatch on cross-architecture transfer while improving robustness to both image- and text-side perturbations, reducing the average degradation $\delta$ on both modalities.
At $N{=}200$, RAHA yields a clear gain in mean transfer (from $7.2$ to $8.7$), with improvements that are consistent across both BERT and DistilBERT targets and across vision backbones.
At $N{=}500$, transfer and robustness become less capacity-limited, and both methods generally improve; importantly, RAHA remains competitive in this higher-budget regime, indicating that the relevance structure distilled by RAHA does not rely on a single encoder geometry and continues to generalize as synthetic capacity grows.
Taken together, transfer improves strongly with N, while robustness does not improve uniformly at the highest budget. These trends are consistent with the intended role of RAHA: distilling relevance in a rank-adaptive shared range while regulating residual interactions, which can improve stability under perturbations and reduce overfitting to a particular encoder configuration.

\section{Auxiliary Discussion}
\begin{figure}[!t]
    \vspace{-3mm}
    \centering
    \includegraphics[width=.6\linewidth,height=3.9cm,trim={1cm 0.7cm 0 0},clip]{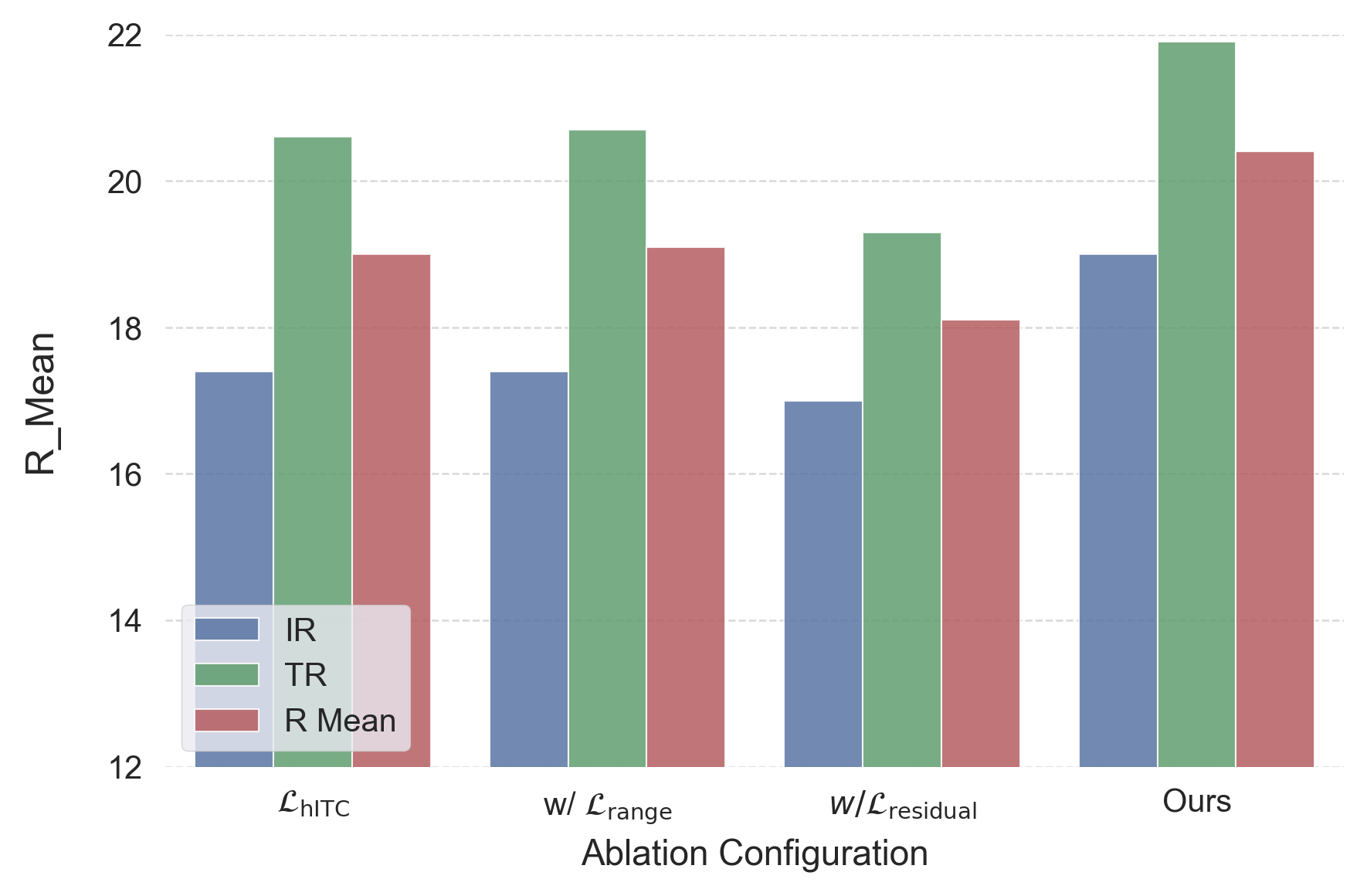}
    \vspace{-3mm}
    \caption{Ablation study for Flickr8k $N{=}100$ setting with each component added, demonstrating the synergy of the two subspace losses.}
    \label{fig:ablation}
    \vspace{-6mm}
\end{figure}

\noindent\textbf{Ablation study.}\quad
Using only hyperbolic contrast $\mathcal{L}_{\mathrm{hITC}}$ provides a strong baseline, confirming that geodesic InfoNCE on synthetic pairs already yields retrieval-relevant alignment (Fig.~\ref{fig:ablation}).
Adding the range relevance term further improves mean retrieval, supporting the role of the range component as the dominant carrier of cross-modal coupling distilled from real data.
Using the residual term alone is weaker, consistent with residual interactions being harder to match reliably without anchoring to the dominant coupled structure.
Combining range and residual relevance matching with their subspace regularizers yields the best performance, indicating that residual matching is most effective when explicitly controlled rather than optimized in isolation.

\noindent\textbf{Distillation cost.}\quad
For completeness, we report a coarse per-iter wall-clock breakdown on an RTX A6000 into (i) data loading, (ii) model init., and (iii) the synthetic update (distillation step), as a complementary metric that contextualizes the computational profile of each objective. With identical data (0.67s) and model (0.3s) initialization, the main difference arises in the distillation step: at a batch size of 1, both are similar at $\approx 25$\,s, while the difference grows with batch size due to lifting and decomposition particularly (\eg,  at 64, $\approx55$\,s vs. $\approx400$\,s), reflecting the stronger batch-scaling cost of RAHA's structure-aware update. We do not use runtime as the primary comparison axis, since our evaluation focuses on retrieval accuracy, transfer, and robustness under fixed synthetic budgets.

In sum, RAHA distills structured image--text relevance through a rank-aware shared subspace. It does not claim uniform dominance in every extreme-compression cell. RAHA is competitive at \texttt{100} pairs and outperforms \cite{lee2025covmatch} at \texttt{200/500}. 
% The \texttt{100}-pair compression rate also varies by dataset: $1.7$\% (\texttt{F8k}) / $0.3$\% (Flickr30k) / $0.8$\textperthousand{} (MS COCO). 
Here, RAHA brings to the surface an important difficulty in VLDD: dataset-scale semantics can be abstracted into decomposed range/residual bases, but a very small synthetic set may not have enough capacity to materialize those semantic modes. At extreme rates, \eg, on Flickr30k/MS COCO, the selected basis can be stable, but too few pairs may not populate the decomposed structure. Thus, MTT-style \cite{xu2024lors} can remain strong at the smallest budgets. As \# pairs  $\uparrow$, RAHA better materializes these modes. Structurally, RAHA uses the conceptual text hierarchy~\cite{desai2023meru, kim2024hype} as an inductive bias for VLDD, since hyperbolic geometry can keep generic concepts close to many descendants while keeping specific instances separable.

\noindent\textbf{On transfer and robustness comparison.}\quad
\cite{zhang2025repblend} builds on the MTT-based \cite{xu2024lors} line by addressing modality collapse,
whereas RAHA follows the distribution-matching line of \cite{lee2025covmatch}, with hyperbolic range--residual relevance modeling in a methodologically complementary, rather than mutually exclusive, direction.

\vspace{-7mm}
\begin{table}[H]
\centering
\scriptsize
\renewcommand{\arraystretch}{1}
\renewcommand{\aboverulesep}{3pt}
\renewcommand{\belowrulesep}{3pt}
\setlength\tabcolsep{5pt}
\fontsize{11pt}{11} \selectfont
\resizebox{.8\linewidth}{!}{
\begin{tabular}{cccccc}
\toprule
\# \textbf{Pairs} & \textbf{Method} & \textbf{Trainable Text Enc.} & \textbf{Cross-arch} (Mean) & \textbf{Image-$\delta$} (Mean) & \textbf{Text-$\delta$} (Mean) \\
\midrule
\multirow{4}{*}{100/200/500}
& LoRS~\cite{xu2024lors}            & \xmark & 10.2 / 10.8 / 11.0 & 0.5 / 0.5 / 0.5  & 0.5 / 0.5 / 0.6  \\[2pt]
& RepBlend~\cite{zhang2025repblend} & \xmark & 8.1 / 8.9 / 10.7 & 0.6 / 0.5 / 0.5  & 0.5 / 0.6 / 0.5 \\[2pt]
\cmidrule{2-6}
& CovMatch~\cite{lee2025covmatch}   & \cmark & 7.3 / 7.2 / 8.7 & 4.6 / 4.2 / 6.4  & 4.6 / 4.0 / 5.1 \\[2pt]
& Ours                              & \cmark &  6.9 / 8.7 / 12.7 &  4.1 / 5.0 / 6.8  & 3.7 / 4.5 / 6.4 \\
\bottomrule
\end{tabular}
}
\end{table}
\vspace{-7mm}

\begin{figure}[!ht]
    \centering
    \includegraphics[width=.7\linewidth]{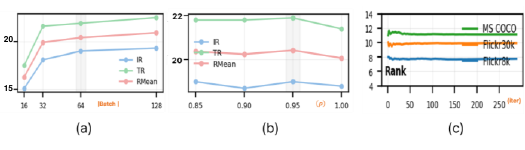}
    \vspace{-5.5mm}
    \caption{Batch, hyperparameter $\rho$ ablations (a,b) and rank by dataset (c).}    
    \label{fig:breakdown}
\end{figure}

\noindent\textbf{Rank selection \& subspace stability.}\quad
RAHA selects the effective rank not manually but based on the smallest $k$ whose cumulative squared singular-value energy of $C_{\mathrm{real}}{=}U\Sigma V^\top$ reaches $\rho$, with numerical-rank clipping and $k_{\max}$ (Eq.11, \S A3.2). This defines the range as the minimal energy-bearing image--text coupling subspace and treats the complement as residual structure. Fig.~\ref{fig:breakdown}(a) shows that larger real batches improve retrieval by stabilizing the cross-covariance estimate, rather than causing $k$ to simply follow the algebraic ceiling $\mathtt{rank}(C_{\mathrm{real}})\leq B{-}1$. Fig.~\ref{fig:breakdown}(b) shows that $\rho{=}0.95$ is the best energy threshold, while $1.0$ degrades by absorbing the low-energy residual tail. Fig.~\ref{fig:breakdown}(c) further shows that the selected rank remains stably within a dataset-specific bound over iterations, supporting stable batch-wise SVD in practice.
\vspace{-8mm}
\vspace{-12mm}
\section{Conclusion} \label{sec:conclusion}

In this work, we present a hyperbolic geometry-aware objective for VLDD, termed \textbf{RAHA}, that lifts image--text representations to \textit{hyperbolic space} and enforces \emph{rank-consistent} cross-modal relevance matching, improving stability and cross-architecture transfer under tight compression budgets.

\noindent\textbf{Limitations.}
As with \cite{wu2024vldistill,xu2024lors,zhang2025repblend,jeong2026mdm,lee2025covmatch}, RAHA remains bounded by the expressivity of chosen teacher encoders, and performance may degrade under domain-shifted or noisy captions. RAHA is driven by inherent hierarchical structures, and thus datasets with weak hierarchy may see limited gains (\S A5). 

\noindent \textbf{Disclosure of LLM Use.}\quad
LLM was used only for editing (\eg, formatting) and did not contribute to the underlying technical ideas or experimental results.

\section*{Acknowledgments}
This work was supported by the Institute of Information \& communications Technology Planning \& Evaluation (IITP) grant funded by the Korea government (MSIT) (No. RS-2024-00457882, AI Research Hub Project) and the National Research Foundation of Korea (NRF) grant funded by the Korea government (MSIT) (RS-2026-25473963).
We thank Jaehyuk Jang for providing key insights.

% \appendix
\title{Rank-Aware Hyperbolic Alignment for Vision–Language Dataset Distillation\\
\textmd{--- Appendix ---}}

\author{
Jongoh Jeong~\inst{1}~\orcidlink{0000-0002-5354-2693},
Sun-Kyung Lee~\inst{2}~\orcidlink{0000-0002-6566-8509}, 
\and
Kuk-Jin Yoon~\inst{1}~\orcidlink{0000-0002-1634-2756} 
}

\authorrunning{Jeong \etal}

\institute{Korea Advanced Institute of Science and Technology (KAIST), Republic of Korea \and
Electronics and Telecommunications Research Institute (ETRI), Republic of Korea\\
\email{\{jeong2, kjyoon\}@kaist.ac.kr, sklee2014@etri.re.kr}
\;\;\href{https://andyj1.github.io/raha}{\textcolor{black}{\small \faGithub\;Project Page}}
}

\titlerunning{RAHA: Appendix}
\maketitle

\setcounter{figure}{0}
\setcounter{table}{0}
\setcounter{equation}{0}
\renewcommand{\thefigure}{A\arabic{figure}}
\renewcommand{\thetable}{A\arabic{table}}
\renewcommand{\theequation}{A\arabic{equation}}
\renewcommand{\thesection}{A\arabic{section}}

\noindent
This appendix provides additional analysis, implementation details, and extended experiments that support the paper, to be added after the main text. Specifically, it covers five parts: the scope of image--text dataset distillation, motivation and additional details of RAHA, experimental details, further analyses and discussion, and broader societal impact as follows:
\begin{enumerate}[leftmargin=1.6em,itemsep=0.25em]
    \item \textbf{Scope of image--text dataset distillation} (Sec.~\S\ref{sec:scope_appendix})
    \item \textbf{Motivation and additional method details} (Sec.~\S\ref{sec:method_appendix})
    \item \textbf{Experimental details} (Sec.~\S\ref{sec:experiments_appendix})
    \item \textbf{Further analyses and discussion} (Sec.~\S\ref{sec:analysis_appendix})
    \item \textbf{Broader societal impact} (Sec.~\S\ref{sec:impact_appendix})
\end{enumerate}

%%%%%%%%%%%%%%%%%%%%%%%%%%%%%%%%%%%%%%%%%%%%%%%%%%%%%%%%%%%%%%%%%%%%%%%%%%%%%%%
\section{Scope of Vision-Language Dataset Distillation}
\label{sec:scope_appendix}

\noindent\fbox{%
\parbox{\dimexpr%
    \linewidth-2\fboxsep-1\fboxrule}%
    {In this section, we clarify the scope of image--text dataset distillation relative to unimodal condensation, and position RAHA as a method for \emph{cross-modal relevance compression} that prioritizes the modality-shared structure.}
}
\vspace{1mm}

\noindent\textbf{Scope.}
Image--text dataset distillation aims to compress a large paired training set into a much smaller synthetic set that remains effective for learning cross-modal models. In our setting, a model is trained from scratch on the distilled set and evaluated primarily by \textbf{bidirectional image-text retrieval}, with additional evaluation on \textbf{prompted image classification} in Sec.~\ref{sec:prompted_cls_eval}. Unlike unimodal dataset distillation, success depends not only on preserving within-modality diversity, but also on retaining the cross-modal alignment structure that determines which captions should be retrieved for an image and which images should be retrieved for a caption. We further examine whether this distilled structure transfers across architectures and remains robust under noise. The goal is therefore not merely dataset compression, but preservation of the shared structure most relevant for retrieval and generalization under a limited synthetic budget.

Existing image--text distillation methods preserve this structure only indirectly. Trajectory-based approaches transfer optimization dynamics from real-data training, which can improve fidelity but require storing long trajectories and do not explicitly identify which components of image--text correspondence matter most. Distribution-based approaches are more storage-efficient, but typically operate in Euclidean feature space and align statistics more uniformly across representation directions. As a result, they may preserve broad feature similarity without sufficiently emphasizing the dominant shared relations most critical for retrieval. Generative approaches provide another route to scalable multimodal compression, but they offer less direct control over how image--text correspondence is preserved in the final distilled set. Across these families, the common limitation is that, under strong compression, not all directions in the joint representation space are equally informative, yet most existing objectives do not explicitly prioritize the most retrieval-relevant shared structure.

\noindent\textbf{Our approach.}
RAHA is designed for this setting. Our starting point is that image--text correspondence is structured rather than isotropic: some directions encode dominant shared relevance, whereas others reflect weaker or noisier residual interactions. To model this more effectively, we redesign the alignment space through a hyperbolic lifting, using geometry as an inductive bias for the non-uniform semantic structure often present in image--text data. We do not assume that paired image--text data form a literal tree; rather, the hyperbolic space provides a more suitable geometry for preserving uneven shared structure than a purely Euclidean formulation. Within this lifted space, RAHA emphasizes the dominant shared component of cross-modal relevance while regulating weaker residual structure, so that limited synthetic capacity is used more effectively for retrieval, transfer, and robustness.

From this perspective, image--text dataset distillation is best viewed as a problem of \textbf{cross-modal relevance compression}. The objective is not to reproduce every aspect of the original multimodal feature distribution, but to preserve the shared structure that is most important for efficient and transferable retrieval learning in the smallest possible synthetic set.

%%%%%%%%%%%%%%%%%%%%%%%%%%%%%%%%%%%%%%%%%%%%%%%%%%%%%%%%%%%%%%%%%%%%%%%%%%%%%%%
\section{Motivation and Additional Method Details}
\label{sec:method_appendix}
\noindent\fbox{%
\parbox{\dimexpr%
    \linewidth-2\fboxsep-2\fboxrule}%
    {In this section, we describe our motivations for the two core design decisions: why RAHA introduces hyperbolic lifting, and why it transfers only the information-rich shared structure.}
}
\vspace{1mm}

\noindent\emph{\textbf{RQ:} Why introduce hyperbolic lifting?}
Image--text semantics often have a hierarchical character~\cite{desai2023meru}. Captions move across levels of abstraction, from scene-level concepts to activities, attributes, and fine-grained object relations. This structure is not naturally flat. Hyperbolic geometry provides a more suitable inductive bias because it can represent coarse-to-fine variation more naturally than Euclidean similarity alone. In RAHA, hyperbolic lifting is not intended to impose a literal tree. Its role is to provide a geometry that better accommodates hierarchical semantic variation.

Hyperbolic lifting is also motivated by the fact that image and text need not coincide even when they are semantically matched. Captions are often more abstract than the images they describe. A geometry that allows structured radial separation is therefore preferable to one that encourages all matched pairs to collapse into a single flat region. This is especially important in distillation, where aggressive collapse can remove useful structure when only a small number of synthetic pairs is available.

\noindent\emph{\textbf{RQ:} Why decompose features in the hyperbolic tangent space?}
Hyperbolic geometry provides the right inductive bias, but decomposition is more stable and interpretable in the tangent space. RAHA therefore performs the decomposition on tangent-space features induced by hyperbolic lifting. This retains the geometry of the lifted representation while allowing dominant and residual directions to be estimated with standard linear operations.

The reason for this decomposition is intuitive. Not all image--text interactions are equally useful for distillation. The strongest shared directions usually carry the most reliable retrieval signal, while the weaker complement often contains low-energy, unstable, or modality-specific structure. Under strong compression, matching all of these interactions equally can waste synthetic capacity and reduce transferability.

\noindent\emph{Logarithmic map at the origin.}
The tangent-space decomposition in \S3.3 relies on the logarithmic map $\mathrm{Log}_o^c$, the inverse of the exponential map defined in Eq.~3.
Given a point on the hyperboloid with spatial component $\bar{u}\in\mathbb{R}^d$, the time coordinate is $u_0=\sqrt{1/c+\|\bar{u}\|_2^2}$.
The logarithmic map at the origin recovers the tangent vector:
\begin{equation}
  \mathrm{Log}_o^c(\bar{u})
  =
  \frac{\operatorname{arcosh}(u_0)}{\sqrt{c}\,\|\bar{u}\|_2}\,\bar{u}
  \;\in\mathbb{R}^d.
  \tag{A.1}
  \label{eq:log_map_def}
\end{equation}
This maps hyperboloid points back to the tangent plane, where Euclidean linear algebra (centering, SVD, projection) can be applied.
The round-trip composition $\mathrm{Log}_o^c \circ \mathrm{Exp}_o^c$ is not the identity: it applies a radial reweighting $f(r)=\operatorname{arcosh}(\cosh(\sqrt{c}\,r))/(\sqrt{c}\,r)$ that compresses large-norm features more than small-norm ones, concentrating the distribution near the origin in a curvature-dependent manner.
In the limit $c\to 0$, this reweighting approaches the identity and the tangent features reduce to the original Euclidean embeddings.

\noindent\emph{\textbf{RQ:} Why prioritize distilling the information-rich shared structure?}
RAHA is built on the intuition that distilled pairs should preserve \emph{useful shared structure}, not full feature correspondence. The dominant shared component carries the most stable image--text relevance information. Matching this component directly encourages the synthetic set to preserve retrieval-oriented structure. In contrast, uniformly matching weak residual directions can amplify noise or modality-specific artifacts that do not help ranking.

This selectivity is most important when the synthetic budget is very small. A compact synthetic set cannot reproduce every degree of freedom in the real data. It should therefore prioritize the dominant shared signal and regulate weaker residual structure. This is the central intuition behind RAHA. Hyperbolic lifting provides a geometry aligned with hierarchical semantics, and tangent-space decomposition provides a practical way to preserve the most retrieval-relevant shared structure.

\noindent\textbf{Relation to prior methods.}
From this perspective, RAHA differs from methods that match Euclidean second-order statistics more uniformly across feature space, and from methods that emphasize low-rank similarity without explicitly separating dominant shared directions from weaker residual ones. The advantage of RAHA is not only its geometry. It is its selective allocation of alignment capacity to the most informative shared structure.

\noindent\textbf{Notation summary.}
For reference throughout the paper and the appendix, we organize and report key notations in Table~\ref{tab:notation} below.

\begin{table}[h]
\centering
\caption{Summary of notations.}
\label{tab:notation}
\vspace{-3mm}
\small
\begin{tabular}{cl}
\toprule
\textbf{Notation} & \textbf{Description} \\
\midrule
$\mathcal{D}_{\mathrm{real}},\; \mathcal{D}_{\mathrm{syn}}$ & Real and synthetic paired datasets \\
$N,\; M$ & Number of real and synthetic pairs \\
$B,\; \tilde{B}$ & Real and synthetic batch sizes \\
$E_v,\; E_t$ & Image and text encoders \\
$\pi_v,\; \pi_t$ & Image and text projection heads \\
$z^v,\; z^t$ & Projected embeddings ($\in\mathbb{R}^d$) \\
$h^v,\; h^t$ & Hyperboloid spatial components after lifting \\
$x^v,\; x^t$ & Tangent-space features after Log map \\
$c$ & Lorentz curvature parameter \\
$s$ & Lifting scale parameter \\
$\tau$ & hITC temperature \\
$\tau_r$ & Relevance temperature \\
$C_{\mathrm{real}},\; C_{\mathrm{syn}}$ & Cross-covariance matrices ($\in\mathbb{R}^{d\times d}$) \\
$U_k,\; V_k$ & Range basis matrices ($\in\mathbb{R}^{d\times k}$) \\
$\rho$ & Energy threshold for SVD rank selection \\
$k$ & Effective coupling rank \\
$a^v,\; a^t$ & Range coordinates ($\in\mathbb{R}^k$) \\
$x^v_{\mathrm{res}},\; x^t_{\mathrm{res}}$ & Residual components  ($\in\mathbb{R}^d$) \\
$G^{\mathrm{range}},\; G^{\mathrm{res}}$ & Range and residual similarity matrices \\
$\Gamma$ & Cost matrix \\
$T$ & Sinkhorn coupling matrix \\
$\varepsilon$ & Sinkhorn entropic regularization \\
$\epsilon$ & Numerical stability constant ($10^{-8}$) \\
$\lambda_{\mathrm{range}},\; \lambda_{\mathrm{residual}},\; \lambda_{\mathrm{comp}}$ & Loss weighting factors  \\
\bottomrule
\end{tabular}
\vspace{-5mm}
\end{table}

%%%%%%%%%%%%%%%%%%%%%%%%%%%%%%%%%%%%%%%%%%%%%%%%%%%%%%%%%%%%%%%%%%%%%%%%%%%%%%%
\section{Experimental Details}
\label{sec:experiments_appendix}
\noindent\fbox{%
\parbox{\dimexpr%
    \linewidth-2\fboxsep-2\fboxrule}%
    {In this section, we provide full implementation details, hyperparameter settings, dataset-selection rationale, extended quantitative comparisons including a reproducibility analysis of the baseline~\cite{lee2025covmatch}, and prompted classification evaluation with their qualitative results.}
}
\vspace{1mm}

\subsection{Workbench Environment}
\label{sec:system_appendix}

All experiments were conducted on a Linux workstation with an Intel Xeon Silver 4210R CPU and a single
NVIDIA RTX A6000 GPU. The software environment used Python 3.10, PyTorch 2.6.0, and Torchvision 0.21.0,
with CUDA 11.8 and cuDNN 9.1.0.
Unless otherwise specified, all compared methods follow the same default training setup, including batch sizes,
optimizer parameters, and evaluation protocol as in \cite{lee2025covmatch} for fair comparisons. 
For RAHA, we use the same settings across datasets and synthetic budgets, unless explicitly stated otherwise.

%%%%%%%%%%%%%%%%%%%%%%%%%%%%%%%%%%%%%%%%%%%%%%%%%%%%%%%%%%%%%%%%%
\subsection{Implementation Details}
\label{sec:impl_appendix}

We follow the standard image--text distillation setup in which synthetic images are optimized directly in pixel space,
while synthetic text is optimized in continuous embedding space. Each distilled query consists of one
$3 \times 224 \times 224$ synthetic image and one 768-dimensional text embedding. The retrieval model uses
an NFNet image encoder and a BERT text encoder, each followed by a trainable projection head into a shared embedding space.

Unless otherwise stated, all experiments follow the distillation-evaluation protocols as in \cite{lee2025covmatch}, including finetuning text encoder layers.
As with existing methods~\cite{wu2024vldistill, xu2024lors, zhang2025repblend, lee2025covmatch}, synthetic images and synthetic text embeddings are updated by an SGD~\cite{bottou2012sgd} optimizer with learning rate 1, respectively, and momentum 0.5. The online retrieval model is also optimized with SGD: the image encoder and text encoder use a learning rate of 0.01, while the image and text projection heads use a learning rate of 0.1. We perform at most 200 distillation iterations, with outer-/inner-loop steps of 50/1, respectively. The distilled set is evaluated every 50 iterations, and each evaluation reports the average over 5 runs. After distillation, a retrieval model is trained from scratch on the distilled set for 100 epochs and evaluated on the real test split.
The default optimization and evaluation settings used throughout the experiments are summarized in Table~\ref{tab:training_hparams}.

\begin{table}[!t]
    \centering
    \caption{Default optimization and evaluation settings used throughout the experiments unless otherwise specified.}
    \label{tab:training_hparams}
    \vspace{-3mm}
    \footnotesize
    \begin{minipage}[t]{0.48\columnwidth}
        \centering
        \setlength{\tabcolsep}{3pt}
        \begin{tabular}{lc}
        \toprule
        \multicolumn{1}{l}{\textbf{Hyperparameter}} & \multicolumn{1}{c}{\textbf{Value}} \\
        \midrule
            Synthetic optimizer & SGD \\
            lr image / text & 1.0 / 1.0 \\
            Synthetic momentum & 0.5 \\
            Online optimizer & SGD \\
            lr enc image / proj & 0.01 / 0.10 \\
            lr enc text / proj & 0.01 / 0.10 \\
            Online momentum & 0.9 \\
            Weight decay & \(5\times10^{-4}\) \\
            Max iterations & 200 \\
            Outer / inner loop iteration & 50 / 1 \\
            Batch size & 64 / 64 \\
            Eval runs & 5 \\
            Eval train epochs & 100 \\
        \bottomrule
        \end{tabular}
    \end{minipage}
\end{table}

\noindent\textbf{RAHA-specific hyperparameters.}
Table~\ref{tab:raha_hparams} lists the RAHA-specific hyperparameters used throughout the experiments.
These values are fixed across all datasets and budgets unless stated otherwise.

\begin{table}[!t]
\centering
\caption{RAHA-specific hyperparameters used in all experiments.}
\label{tab:raha_hparams}
\vspace{-3mm}
\begin{tabular}{lcc}
\toprule
\textbf{Hyperparameter} & \textbf{Notation} & \textbf{Value} \\
\midrule
Lorentz curvature & $c$ & 1.0 \\
Lifting scale & $s$ & 1.0 \\
hITC temperature & $\tau$ & 0.07 \\
Relevance temperature & $\tau_r$ & 0.07 \\
Energy threshold (rank selection) & $\rho$ & 0.95 \\
Sinkhorn regularization & $\varepsilon$ & 0.05 \\
Sinkhorn iterations & -- & 20 \\
Numerical stability constant & $\epsilon$ & $10^{-8}$ \\
\midrule
Range loss weight & $\lambda_{\mathrm{range}}$ & 0.8 \\
Residual loss weight & $\lambda_{\mathrm{residual}}$ & 0.4 \\
Compression weight (within residual) & $\lambda_{\mathrm{comp}}$ & 0.1 \\
\bottomrule
\end{tabular}
\end{table}

In particular, we distinguish two small constants used in the paper, $\varepsilon$ and $\epsilon$.
The symbol $\varepsilon=0.05$ denotes the entropic regularization strength in the Sinkhorn transport solver (Eq.~16), which controls the smoothness of the coupling matrix $T$.
The symbol $\epsilon=10^{-8}$ is a numerical stability floor used in denominators and clamping operations.

\noindent\textbf{Scope of baselines.}
RepBlend~\cite{zhang2025repblend} operates within the trajectory-matching paradigm, inheriting expert-trajectory construction and LoRS-style low-rank similarity learning.  It is therefore not a direct counterpart to distribution-matching methods such as CovMatch or RAHA.  We include LoRS~\cite{xu2024lors} as the representative trajectory-based baseline with explicit low-rank modeling.  RepBlend's contributions are orthogonal and could in principle be combined with distribution-matching frameworks.

\noindent\textbf{Synthetic data initialization.}
By default, we initialize the synthetic set by sampling $M$ image--text pairs uniformly at random from $\mathcal{D}_{\mathrm{real}}$.
Each synthetic image $\tilde{x}_j$ is initialized as the pixel tensor of a randomly selected real image, and each synthetic text embedding $\tilde{\mathbf{E}}_j$ is initialized by passing the corresponding real caption through the text encoder's \textit{embedding layer} including positional encoding, before the Transformer encoder layers.
% The attention mask $\tilde{\mathbf{m}}_j$ is set by the tokenizer at initialization and remains fixed throughout optimization.
% We also support coreset-based initialization (herding, $k$-center greedy, forgetting-based selection), where the $M$ pairs are selected by a coreset method before continuous optimization begins.
% All results in the main paper use random initialization unless stated otherwise.

\noindent\textbf{Model initialization.}
At the start of each distillation iteration, the encoder $\Psi$ is reset to its pretrained weights before the outer loop begins.
This follows the offline-online training protocol practiced in CovMatch~\cite{lee2025covmatch} and prevents the encoder from drifting into a configuration that overfits the current synthetic set. Without this per-iteration reset, the encoders can adapt to exploit synthetic-specific artifacts (\eg, high-frequency texture patterns), potentially yielding a distilled set with poor transferability when a fresh encoder is trained from scratch during evaluation.
The reset ensures that the distillation gradient is computed under a diverse set of encoder states (as the inner-loop updates produce a slightly different trajectory each iteration), which encourages the synthetic data to generalize across encoder configurations.

\noindent\textbf{Numerical stability of Singular Value Decomposition (SVD).}
The tangent-space cross-covariance $C_{\mathrm{real}}$ can be ill-conditioned, particularly early in training or with small batch sizes.
To ensure stable rank selection, we apply a jittered SVD by computing $\mathrm{SVD}(C_{\mathrm{real}} + \delta I)$ with
$\delta = 10^{-6}\cdot\|C_{\mathrm{real}}\|_F / d$ and $I$ being a rectangular identity of matching dimensions, using \texttt{torch.linalg.svd}.
In case of SVD computation failure due to ill-conditioning, we alternatively compute
the SVD indirectly via the eigendecomposition of $C_{\mathrm{real}}C_{\mathrm{real}}^\top$ and $C_{\mathrm{real}}^\top C_{\mathrm{real}}$, symmetrized to avoid numerical asymmetry.

\noindent\textbf{Sinkhorn cost ($\Gamma$) normalization.}
Before computing the Gibbs kernel $K=\exp(-\Gamma/\varepsilon)$, we normalize the cost matrix by its mean: $\Gamma \leftarrow \Gamma / \bar{\Gamma}$, where $\bar{\Gamma} = \frac{1}{\tilde{B}B}\sum_{ij}\Gamma_{ij}$.
This prevents kernel entries from collapsing to zero when absolute cost values are large relative to $\varepsilon$, and makes the effective regularization strength approximately invariant to the scale of KL divergences across different training stages.

%%%%%%%%%%%%%%%%%%%%%%%%%%%%%%%%%%%%%%%%%%%%%%%%%%%%%%%%%%%%%%%%%
\subsection{Deliberate Selection of Datasets by Scale}
\label{sec:datasets_appendix}

We evaluate on Flickr8k, Flickr30k, and COCO using the standard Karpathy-style retrieval splits. These datasets cover a useful range of scales (refer to Table {\textcolor{red}{3}} in the main paper), from a relatively small and clean benchmark to substantially larger and more diverse image--text collections. This choice allows to gauge whether the same distillation principle remains effective across dataset scale and diversity. Flickr8k provides a small-scale setting where coverage is limited and overfitting is a practical concern. Flickr30k is a standard mid-scale benchmark. COCO is larger and more diverse, with broader compositional variation. Together, these datasets provide a useful picture of how each method behaves as the source distribution becomes more complex.

This evaluation is particularly informative for RAHA, whose advantage is expected to grow as synthetic capacity increases. The results in the main paper confirm this trend: RAHA remains competitive at small budgets and improves steadily as~$N$ grows.

\noindent\textbf{Sensitivity to real Euclidean cross-covariance.}
CovMatch is a strong Euclidean statistics-matching baseline, but its behavior depends on how strongly the real Euclidean cross-covariance target is enforced during distillation. In its released implementation, the covariance term is written as $\|C_{\mathrm{syn}}-\rho C_{\mathrm{real}}\|_F^2$, while $\lambda$ controls the relative strength of auxiliary feature-matching losses. Thus, $\rho$ rescales the real cross-covariance target, whereas $\lambda$ controls the balance between covariance alignment and feature-level regularization. In CovMatch, they report different $(\rho,\lambda)$ settings across budgets, which indicates that its performance is sensitive to this balance.

In contrast, RAHA does \textbf{not} rely on rescaling the real cross-covariance target across settings. Instead, it imposes structured supervision through an explicit decomposition into range and residual subspaces, so that dominant shared structure and weaker residual structure are treated differently by design. This reduces dependence on 
pair-level tuning and aligns more naturally with our geometric motivation.

%%%%%%%%%%%%%%%%%%%%%%%%%%%%%%%%%%%%%%%%%%%%%%%%%%%%%%%%%%%%%%%%%
\subsection{Extended Quantitative Results}
\label{sec:quantitative_appendix}

\noindent\textbf{On Reproducibility.} 
% (Tables~\ref{tab:covmatch_reimpl}~and~\ref{tab:repro_gap}).
(Table~\ref{tab:repro_gap}).
\label{sec:reproducibility}
Exact reproduction of CovMatch from the released public artifacts is not straightforward.  The public repository provides script entry points for multiple dataset--budget settings, but the released hyperparameters are not fully synchronized with the values reported in the paper.  Specifically, the public code uses $\rho = 2$ for the 200-pair setting where the paper reports $\rho = 1$, and $\lambda = 0.5$ for the 500-pair setting where the paper reports $\lambda = 0.6$.  The code repository indicates that only the Flickr30k 100-pair synthetic set is directly released; other configurations must be re-run from scratch.

Table~\ref{tab:repro_gap} summarizes the published-to-reproduced gap alongside RAHA's performance at each budget.
Under the reproduced protocol, which we consider the fairest publicly verifiable comparison, RAHA outperforms CovMatch at 200 and 500 pairs on all three datasets and remains competitive at 100 pairs.  This pattern is consistent across Flickr8k (where the published and reproduced CovMatch numbers are closer), Flickr30k, and COCO.  
% We therefore present both comparisons: Table~4 in the main paper preserves the published CovMatch results for direct comparability with the original paper.

\noindent\textbf{Higher-scale comparison on CC3M-595K-LLaVA.}
% \noindent\textbf{Benchmark dataset-dependent phenomenon.}
The relatively smaller gains on Flickr30k and COCO at low budgets reflect properties of these benchmarks rather than a fundamental limitation of hyperbolic geometry.  Both datasets contain multiple captions per image with substantial semantic overlap, creating a regime in which precise instance discrimination can dominate over global hierarchy recovery.  This can reduce the visible benefit of a hierarchy-aware geometry even when the underlying structure remains present.  The trend reverses at larger budgets,
% (Tables~\ref{tab:covmatch_reimpl},~\ref{tab:extended_1000pairs}),
 where RAHA's advantage becomes clear across all three datasets. 

In this light, we complement the Flickr30k and COCO evaluations with an additional evaluation in Table~\ref{tab:cc3m_compare} on CC3M-595K-LLaVA~\cite{llava_cc3m_595k}, a \textit{subset} of CC3M due to the inaccessibility of the CC3M dataset in full from the original source. \textit{CC3M-595K-LLaVA (595k in scale) stands as a broader and noisier image--text source than the standard retrieval benchmarks, providing a useful stress test at larger scale.}
At 100 pairs, RAHA matches CovMatch as with the Flickr8k case. At 200 pairs, RAHA tops CovMatch. At 500 pairs, the gain becomes substantial, with the mean score improving from 5.1 to 8.1. These results indicate that the selective preservation of information-rich shared structure becomes increasingly beneficial as the source distribution becomes broader and the synthetic budget becomes sufficiently expressive.

\begin{table}[!t]
  \centering
  \caption{Gap between published and reproduced 
    % {CovMatch\textcolor{red}{*}}
    CovMatch mean retrieval scores for Flickr30k and COCO.  $\Delta$ denotes the shortfall of the
    reproduced result relative to the published number.  Under the
    reproducible protocol ($|B|=64$), RAHA outperforms CovMatch at 200 and 500 pairs
    on both datasets and remains competitive at 100 pairs.}% \textbf{We will supplement this in Table~{\textcolor{red}{4}}.}}
  \label{tab:repro_gap}
  \vspace{-3mm}
  \setlength{\tabcolsep}{5pt}
  \footnotesize
  \begin{tabular}{llcccc}
    \toprule
    \textbf{Dataset} & \textbf{\# Pairs}
      & \textbf{Published} & \textbf{Reproduced} %{\textcolor{red}{*}}
      & $\boldsymbol{\Delta}$ & \textbf{RAHA} \\
    \midrule
    \multirow{3}{*}{Flickr30k}
      & 100 & 30.5 & 22.8 & $-7.7$  & 20.7 \\
      & 200 & 34.4 & 22.0 & $-12.4$ & \textbf{25.7} \\
      & 500 & 38.4 & 28.9 & $-9.5$  & \textbf{32.9} \\
    \midrule
    \multirow{3}{*}{COCO}
      & 100 & 11.5 & 7.0  & $-4.5$ & \textbf{7.2} \\
      & 200 & 14.8 & 8.3  & $-6.5$ & \textbf{10.2} \\
      & 500 & 19.6 & 11.2 & $-8.4$ & \textbf{13.7} \\
    \bottomrule
  \end{tabular}
\end{table}

\begin{table}[!t]
\centering
\footnotesize
\caption{Comparison on CC3M-595K-LLaVA across 100, 200, and 500 pairs. IR and TR denote mean image-retrieval and text-retrieval scores averaged over K@\{1,5,10\}, and Mean is their average. RAHA matches CovMatch at 100 pairs and becomes clearly stronger as the synthetic budget increases.}
\label{tab:cc3m_compare}
\vspace{-3mm}
\setlength{\tabcolsep}{4pt}
\begin{tabular}{lccccc}
\toprule
\textbf{Dataset} & \textbf{\# Pairs} & \textbf{Method} & \textbf{IR} & \textbf{TR} & \textbf{Mean} \\
\midrule
\multirow{2}{*}{CC3M-595K-LLaVA~\cite{llava_cc3m_595k}} & \multirow{2}{*}{100} & CovMatch~\cite{lee2025covmatch} & 3.6 & 4.3 & 4.0 \\
& & \cellcolor{eccvblue!20}Ours    & \cellcolor{eccvblue!20}3.7 & \cellcolor{eccvblue!20}4.3 & \cellcolor{eccvblue!20}4.0 \\
\midrule
\multirow{2}{*}{CC3M-595K-LLaVA~\cite{llava_cc3m_595k}} & \multirow{2}{*}{200} & CovMatch~\cite{lee2025covmatch} & 3.9 & 4.3 & 4.1 \\
&  & \cellcolor{eccvblue!20}Ours     & \cellcolor{eccvblue!20}5.1 & \cellcolor{eccvblue!20}5.7 & \cellcolor{eccvblue!20}5.4 \\
\midrule
\multirow{2}{*}{CC3M-595K-LLaVA~\cite{llava_cc3m_595k}} & \multirow{2}{*}{500} & CovMatch~\cite{lee2025covmatch} & 4.8 & 5.4 & 5.1 \\
 &  & \cellcolor{eccvblue!20}Ours     & \cellcolor{eccvblue!20}7.6 & \cellcolor{eccvblue!20}8.6 & \cellcolor{eccvblue!20}8.1 \\
\bottomrule
\end{tabular}
\end{table}

\noindent\textbf{Larger-budget comparison.}
Table~\ref{tab:extended_1000pairs} reports the 1000-pair comparison. RAHA improves clearly over CovMatch on Flickr8k and Flickr30k in mean retrieval. On COCO, RAHA also improves the mean score over CovMatch. These results show that the advantage of geometry-aware relevance distillation becomes more pronounced once the synthetic set has enough capacity to retain structured shared information.

\begin{table}[!t]
\centering
\footnotesize
\caption{Extended 500/1000-pair image-text retrieval comparison. IR and TR denote mean image-retrieval and text-retrieval scores, and Mean is their average. RAHA improves over 
% {CovMatch\textcolor{red}{*}}
CovMatch on all three datasets at this larger synthetic budget.} %``-'' indicates N/A, and {\textcolor{red}{*}} denotes reproduced.}
\label{tab:extended_1000pairs}
\vspace{-3mm}
\setlength{\tabcolsep}{4pt}
    \begin{tabular}{lcccccc}
    \toprule
        \textbf{Dataset} & \textbf{\# Pairs} & \textbf{Method} & \textbf{IR} & \textbf{TR} & \textbf{Mean} \\
    \midrule
        \multirow{6}{*}{Flickr8k~\cite{hodosh2013flickr8k}} & \multirow{3}{*}{500}  & EDGE~\cite{zhao2025edge} & \textcolor{gray}{\texttt{N/A}} & \textcolor{gray}{\texttt{N/A}} & \textcolor{gray}{\texttt{N/A}}\\
        &  & CovMatch~\cite{lee2025covmatch} & 23.8 & 28.0 & 25.9 \\
        &  & \cellcolor{eccvblue!20}Ours     & \cellcolor{eccvblue!20}27.7 & \cellcolor{eccvblue!20}33.6 & \cellcolor{eccvblue!20}30.7 \\
    \cmidrule{2-6}
         & \multirow{3}{*}{1000}  & EDGE~\cite{zhao2025edge}  & \textcolor{gray}{\texttt{N/A}} & \textcolor{gray}{\texttt{N/A}} & \textcolor{gray}{\texttt{N/A}}\\
         & & CovMatch~\cite{lee2025covmatch} & 28.0 & 33.2 & 30.6 \\
        &  & \cellcolor{eccvblue!20}Ours     & \cellcolor{eccvblue!20}33.6 & \cellcolor{eccvblue!20}40.7 & \cellcolor{eccvblue!20}37.1 \\
    \midrule
        \multirow{6}{*}{Flickr30k~\cite{young2014flickr}}& \multirow{3}{*}{500} & EDGE~\cite{zhao2025edge} & 19.4 & 32.1 & 25.8 \\
        & & 
        % {CovMatch
        % \textcolor{red}{*}
        % }
        CovMatch~\cite{lee2025covmatch} & 26.3 & 31.5 & 28.9 \\
        & & \cellcolor{eccvblue!20}Ours     & \cellcolor{eccvblue!20}30.0 & \cellcolor{eccvblue!20}35.9 & \cellcolor{eccvblue!20}32.9 \\
    \cmidrule{2-6}
        & \multirow{3}{*}{1000} & EDGE~\cite{zhao2025edge}  & 26.2 & 34.8 & 30.5\\
        & & CovMatch~\cite{lee2025covmatch} & 26.5 & 30.2 & 28.4 \\
        & & \cellcolor{eccvblue!20}Ours     & \cellcolor{eccvblue!20}34.5 & \cellcolor{eccvblue!20}41.5 & \cellcolor{eccvblue!20}38.0 \\
    \midrule
        \multirow{6}{*}{COCO~\cite{COCO}}& \multirow{3}{*}{500}      & EDGE~\cite{zhao2025edge} & 6.5 & 9.4 & 7.9 \\
        & & 
        % {CovMatch
        %
        % \textcolor{red}{*}
        % }
        CovMatch~\cite{lee2025covmatch} & 9.9 & 12.6 & 11.2\\
        &      & \cellcolor{eccvblue!20}Ours     & \cellcolor{eccvblue!20}12.6 & \cellcolor{eccvblue!20}14.9 & \cellcolor{eccvblue!20}13.7 \\
    \cmidrule{2-6}
        & \multirow{3}{*}{1000}      & EDGE~\cite{zhao2025edge}  & 9.6 & 12.6 & 11.1\\
        & & CovMatch~\cite{lee2025covmatch} & 9.6  & 22.1 & 14.2 \\
        &      & \cellcolor{eccvblue!20}Ours     & \cellcolor{eccvblue!20}16.4 & \cellcolor{eccvblue!20}20.8 & \cellcolor{eccvblue!20}18.6 \\
    \bottomrule
    \end{tabular}
\end{table}

\begin{table}[!t]
    \centering
    \caption{\textbf{Prompted image classification results} on CIFAR-100~\cite{cifar} (32$\times$32), CUB-200-2011~\cite{welinder2010cub2002011}, Stanford Cars~\cite{krause2013stanfordcars}, and ImageNet-1K~\cite{imagenet} (224$\times$224). We use the dataset-specific prompts as used in CLIP~\cite{radford2021clip}.}
    \label{tab:prompted_classification}
    \vspace{-3mm}
    \begin{tabular}{cccccc}
     \toprule
        \textbf{Dataset} & \textbf{Image Resolution} & \textbf{\# Pairs} & \textbf{Method} & \textbf{Top-1 Acc.} (\%) \\
    \midrule
         \multirow{2}{*}{CIFAR-100~\cite{cifar}} & \multirow{2}{*}{32$\times$32} & \multirow{2}{*}{100} & CovMatch~\cite{lee2025covmatch} &  14.55\\
						           & && \cellcolor{eccvblue!20}Ours & \cellcolor{eccvblue!20}15.60 \\
    \midrule
        \multirow{2}{*}{CUB-200-2011~\cite{welinder2010cub2002011}} & \multirow{6}{*}{224$\times$224}  & \multirow{6}{*}{100} & CovMatch~\cite{lee2025covmatch} & 12.53 \\
        							  &  && \cellcolor{eccvblue!20}Ours & \cellcolor{eccvblue!20}14.43 \\
    \cmidrule{1-1}\cmidrule{4-5}
        \multirow{2}{*}{Stanford Cars~\cite{krause2013stanfordcars}} & & & CovMatch~\cite{lee2025covmatch} & 5.63 \\
        							   &  && \cellcolor{eccvblue!20}Ours & \cellcolor{eccvblue!20}7.80 \\
   \cmidrule{1-1}\cmidrule{4-5}
        \multirow{2}{*}{ImageNet-1K~\cite{imagenet}} &  &  & CovMatch~\cite{lee2025covmatch} &  7.62 \\
        						       &  && \cellcolor{eccvblue!20}Ours & \cellcolor{eccvblue!20}7.88 \\
    \bottomrule
    \end{tabular}
\end{table}

\noindent\textbf{Comparison with EDGE (Table~\ref{tab:extended_1000pairs}).}
EDGE~\cite{zhao2025edge} reports larger-budget results beyond the standard low-budget setting, making it a useful point of comparison at higher synthetic capacity. This comparison is complementary to CovMatch because the two baselines reflect different design choices. CovMatch~\cite{lee2025covmatch} emphasizes Euclidean statistics matching with a trainable text encoder, whereas EDGE is more generation-oriented by employing Stable Diffusion (SD) v1.5~\cite{rombach2022SD}. RAHA differs from both by focusing on the selective preservation of retrieval-relevant shared structure through geometry-aware matching without an explicitly pretrained generative model like SD.
We note that EDGE operates in a generative paradigm (latent diffusion synthesis of images plus discrete caption generation) and uses a fundamentally different data representation than RAHA, making the comparison complementary rather than strictly head-to-head.
The comparison in Table~\ref{tab:extended_1000pairs} at the budgets where EDGE numbers are available shows that RAHA outperforms EDGE on all three datasets at both 500 and 1000 pairs.

%%%%%%%%%%%%%%%%%%%%%%%%%%%%%%%%%%%%%%%%%%%%%%%%%%%%%%%%%%%%%%%%%
\subsection{Evaluation on Prompted Classification}
\label{sec:prompted_cls_eval}

\begin{figure}[!t]
    \centering
    \includegraphics[width=.99\linewidth]{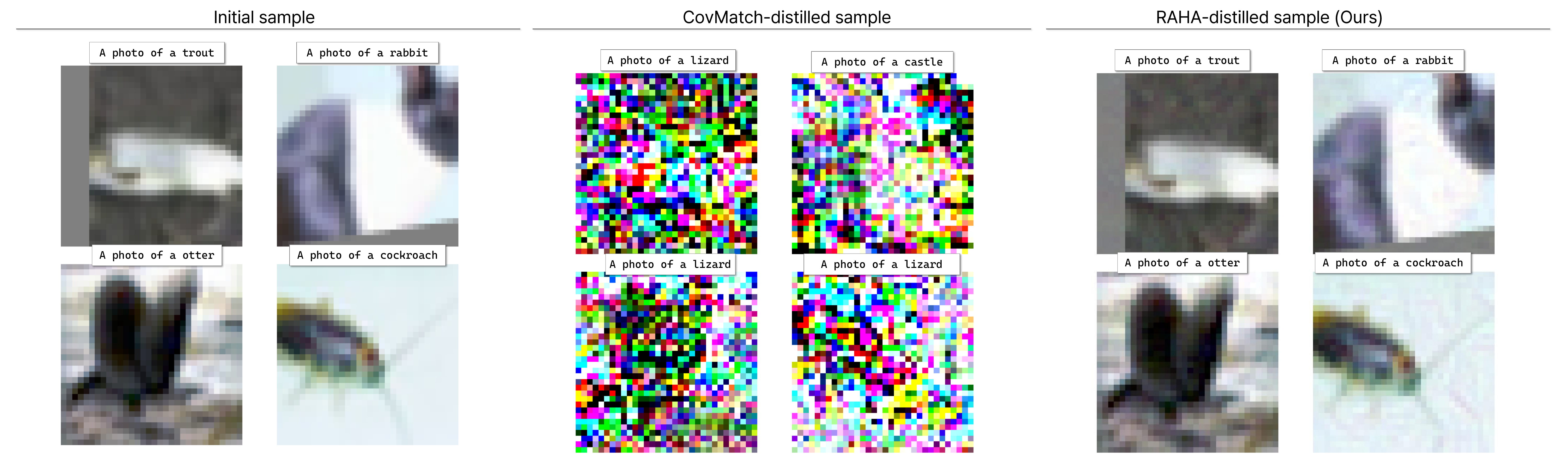}
    \vspace{-3mm}
    \caption{Distilled data comparison on CIFAR-100. Left: initial samples. Middle: CovMatch-distilled~\cite{lee2025covmatch}. Right: RAHA-distilled (Ours). RAHA preserves visual fidelity and image--text consistency more reliably across samples.}
    \label{fig:additional_cls_qual1}
    \vspace{-4mm}
\end{figure}
\begin{figure}[!t]
    \centering
    \includegraphics[width=.99\linewidth]{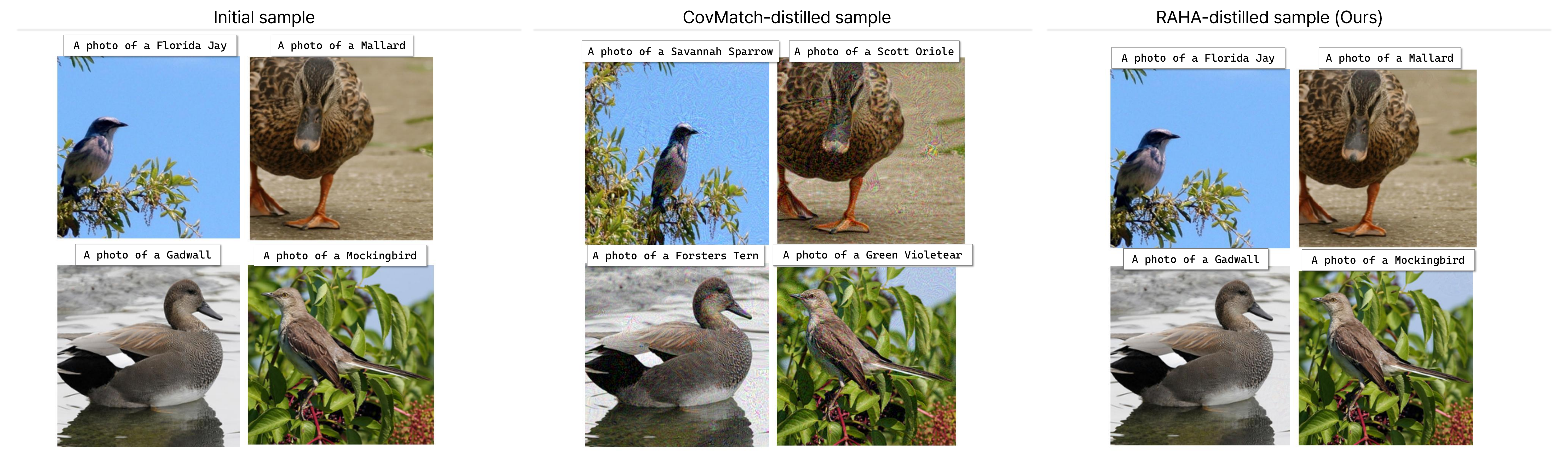}
    \vspace{-3mm}
    \caption{Distilled data comparison on CUB-200-2011. Left: initial samples. Middle: CovMatch-distilled~\cite{lee2025covmatch}. Right: RAHA-distilled (Ours).}
    \label{fig:additional_cls_qual2}
    \vspace{-4mm}
\end{figure}
\begin{figure}[!t]
    \centering
    \includegraphics[width=.99\linewidth]{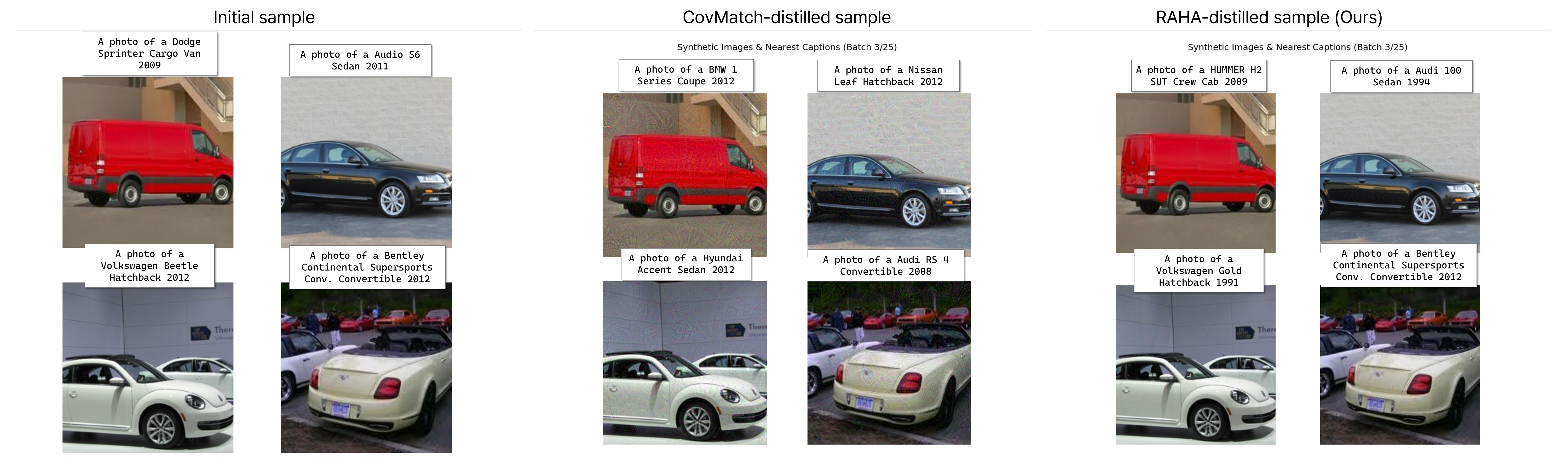}
    \vspace{-3mm}
    \caption{Distilled data comparison on Stanford Cars. Left: initial samples. Middle: CovMatch-distilled~\cite{lee2025covmatch}. Right: RAHA-distilled (Ours).}
    \label{fig:additional_cls_qual3}
    \vspace{-4mm}
\end{figure}
\begin{figure}[!t]
    \centering
    \includegraphics[width=.99\linewidth]{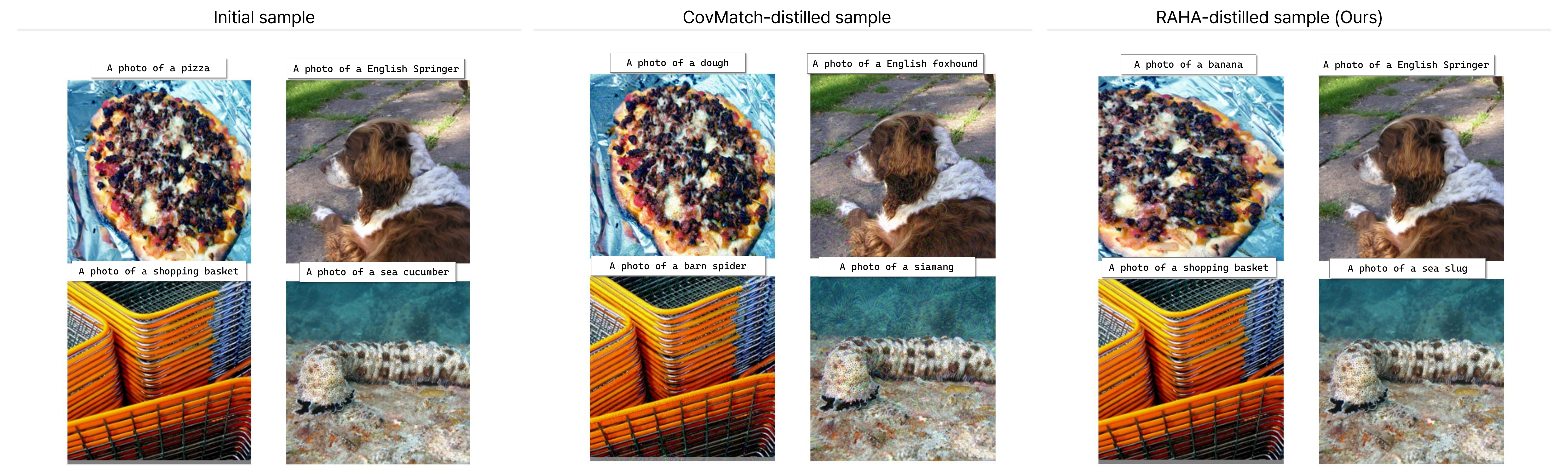}
    \vspace{-3mm}
    \caption{Distilled data comparison on ImageNet-1K. Left: initial samples. Middle: CovMatch-distilled~\cite{lee2025covmatch}. Right: RAHA-distilled (Ours).}
    \label{fig:additional_cls_qual4}
\end{figure}

\noindent\textbf{Prompted image classification task results.}
To assess whether RAHA's distilled pairs preserve structure useful beyond retrieval, we evaluate prompted zero-shot classification on four benchmarks following the CLIP prompting technique~\cite{radford2021clip}.  Table~\ref{tab:prompted_classification} reports Top-1 accuracy at $N{=}100$.  RAHA improves over CovMatch on CIFAR-100 ($+$1.05\,pp), CUB-200-2011 ($+$1.90\,pp), and Stanford Cars ($+$2.17\,pp), and matches on ImageNet-1K ($+$0.26\,pp).  The gains are largest on fine-grained benchmarks (CUB, Cars), where the hierarchical inductive bias of hyperbolic geometry is expected to be most beneficial, as these datasets exhibit taxonomic structure that aligns naturally with coarse-to-fine semantic organization.

%%%%%%%%%%%%%%%%%%%%%%%%%%%%%%%%%%%%%%%%%%%%%%%%%%%%%%%%%%%%%%%%%
\subsection{Additional Qualitative Results}
\label{sec:qualitative_appendix}
The qualitative comparison complements the retrieval tables by showing how distilled image--text pairs evolve during optimization and how well different methods preserve cross-modal consistency. Relative to the strongest Euclidean statistics-based baseline, RAHA tends to produce cleaner visual structure and better preserve fine-grained image--text agreement. This is consistent with our design goal: preserving dominant shared structure while controlling weaker residual interactions.

Here, we additionally provide qualitative comparison of 100-pair distilled data on CIFAR-100, CUB-200-2011, Stanford Cars, and ImageNet datasets between the initial sample, CovMatch~\cite{lee2025covmatch}-distilled and RAHA-distilled synthetic samples in Figs.~\ref{fig:additional_cls_qual1},~\ref{fig:additional_cls_qual2},~\ref{fig:additional_cls_qual3}, and~\ref{fig:additional_cls_qual4}. Broadly, we observe a similar pattern to Fig.~1 of the main paper: CovMatch leaves a significant portion of trailing noise-like artifacts in various regions of the distilled images, with occasional mismatched attributes in the distilled text. By contrast, RAHA-distilled samples produce substantially more realistic distilled data that are geometrically consistent (without noise-like artifacts) and match prompted label text with the initialization while achieving higher classification accuracy than the baseline. 

Interestingly, on CIFAR-100 containing coarse 32$\times$32 resolution images, CovMatch c\textit{ompletely diverges from being realistic in visuals}. Moreover, the text descriptions (labels) are far from the initialization, suggesting that the second-order statistics matching mechanism in CovMatch~\cite{lee2025covmatch} fails to optimize distilled data toward realism. In contrast, our method largely maintains realistic distillation of the real data without undesirable visual artifacts or semantic perturbations in the text.

%%%%%%%%%%%%%%%%%%%%%%%%%%%%%%%%%%%%%%%%%%%%%%%%%%%%%%%%%%%%%%%%%%%%%%%%%%%%%%%
\section{Further Analyses and Discussion}
\label{sec:analysis_appendix}
\noindent\fbox{%
\parbox{\dimexpr%
    \linewidth-2\fboxsep-2\fboxrule}%
    {In this section, we further discuss sensitivity to the hyperparameter, compute profile with a contextualized cost defense, 
    % training dynamics of the range--residual decomposition, 
    and performance at small budgets accompanied by the reproducibility analysis.}
}
\vspace{1mm}

\begin{figure}[!ht]
    \centering
    \includegraphics[width=\linewidth]{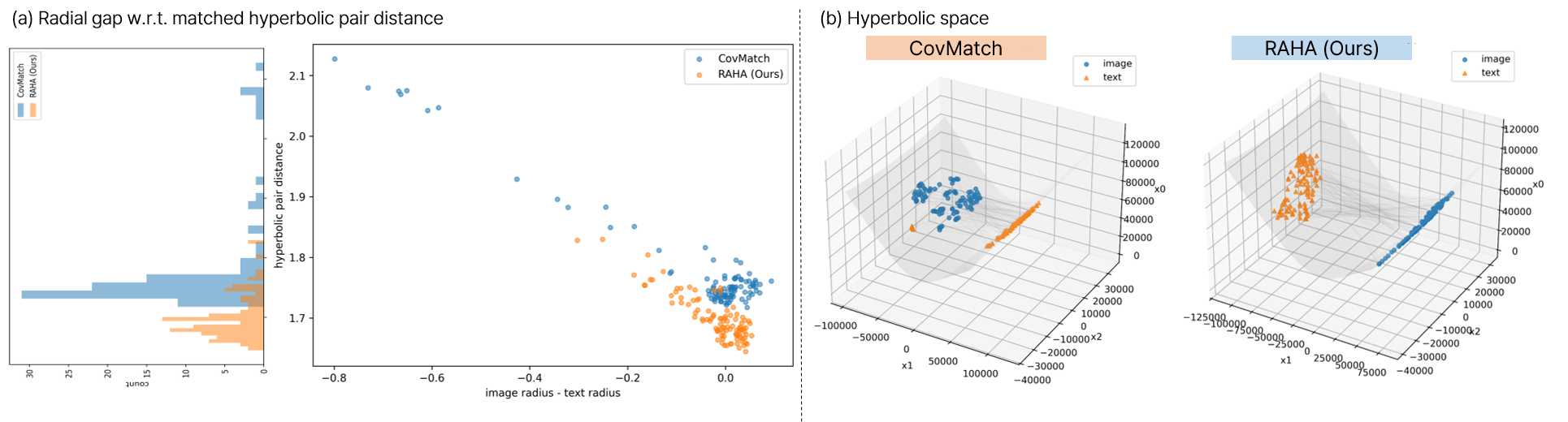}
    \caption{
    (a) Radial gap and matched hyperbolic pair distance. The radial gap
    measures how differently a matched image and text pair are placed with respect to the hyperbolic origin. RAHA yields a tighter radial-gap distribution and lower matched hyperbolic distances than CovMatch, indicating more consistent cross-modal placement at similar hierarchical depth.
    (b) Euclidean and Lorentz views of the distilled features. Compared with CovMatch, RAHA yields a more compatible global image--text configuration in hyperbolic space, with more stable radial ordering and less irregular modality-dependent displacement, consistent with preserving hierarchy-aware cross-modal relevance.
    }
    \label{fig:feature_viz}
\end{figure}

\noindent\textbf{Feature-association visualization.}
We also include a feature visualization in Euclidean and Lorentz spaces to support the geometric motivation of RAHA (Fig.~\ref{fig:feature_viz}). The goal is not to claim exact tree reconstruction. Instead, the visualization tests whether the distilled image and text features retain a more structured relative organization. We focus on three properties: cleaner grouping of semantically related pairs, more stable coarse-to-fine organization, and reduced overlap caused by noisy residual interactions. 

Fig.~\ref{fig:feature_viz}(a) is particularly informative because it separates \emph{hierarchical-depth mismatch} from total cross-modal discrepancy. For each matched pair of Lorentz-lifted points \((x_i^{\mathrm{img}}, x_i^{\mathrm{txt}})\), we plot the radial gap
\begin{align*}
\Delta r_i = r(x_i^{\mathrm{img}}) - r(x_i^{\mathrm{txt}})
\end{align*}
against the matched hyperbolic distance
% \begin{align*}
% d_{\mathcal{H}}\!\big(x_i^{\mathrm{img}},x_i^{\mathrm{txt}}\big)
% =
% \frac{1}{\sqrt{c}}
% \operatorname{arcosh}\!\Big(-c\,\linner{x_i^{\mathrm{img}}}{x_i^{\mathrm{txt}}}\Big).
% \end{align*}
\begin{align*}
d_{\mathcal{H}}\!\big(x_i^{\mathrm{img}},x_i^{\mathrm{txt}}\big)
=
\frac{1}{\sqrt{c}}\,
\operatorname{arcosh}\!\Big(
\max\!\big(
-c\,\linner{x_i^{\mathrm{img}}}{x_i^{\mathrm{txt}}},\;
1+\varepsilon
\big)
\Big).
\end{align*}
In this pair distance formulation, \(r(x)=d_{\mathcal{H}}(o,x)\) is the geodesic radius from the origin, so \(\Delta r_i\) measures whether the paired image and text embeddings are placed at comparable semantic depth in hyperbolic space. Since radius itself is a distance from the origin, the triangle inequality gives:
\begin{align*}
|\Delta r_i|
=
|d_{\mathcal{H}}(o,x_i^{\mathrm{img}})-d_{\mathcal{H}}(o,x_i^{\mathrm{txt}})|
\;\le\;
d_{\mathcal{H}}(x_i^{\mathrm{img}},x_i^{\mathrm{txt}}),
\end{align*}
which means that large radial mismatch necessarily induces a large matched pair distance, regardless of any angular agreement. Thus, panel~(a) is not merely descriptive: it directly diagnoses whether the method places matched image--text pairs at compatible hierarchical levels.

The plot shows that CovMatch exhibits a broader and more negatively shifted distribution of \(\Delta r_i\), indicating that text features are often pushed farther from the origin than their matched image features. These pairs also populate the higher-distance regime, revealing that substantial radial mismatch is coupled with poorer cross-modal association. In contrast, RAHA concentrates pairs much closer to \(\Delta r_i \approx 0\) and simultaneously in a lower \(d_{\mathcal{H}}\) regime. This means that RAHA reduces modality-dependent radial drift and places matched image--text pairs at more compatible semantic depths, not only making them closer in hyperbolic space, but also organizing them more consistently with the intended coarse-to-fine hierarchy. We view this as direct geometric evidence that the rank-aware range--residual supervision preserves hierarchy-aware cross-modal relevance more faithfully than uniform Euclidean-style alignment.

Fig.~\ref{fig:feature_viz}(b) provides a complementary manifold-level view of the distilled features in the
Lorentz model. In contrast to (a) quantifying matched-pair consistency through
\(\Delta r_i\) and \(d_{\mathcal{H}}(x_i^{\mathrm{img}},x_i^{\mathrm{txt}})\),
(b) shows whether these local improvements translate into a more \textit{coherent global
cross-modal geometry}. In CovMatch, the image and text embeddings appear more unevenly
organized, with a less compatible radial arrangement and stronger modality-dependent
displacement. In RAHA, the two modalities exhibit a more consistent joint configuration:
their placements are more geometrically compatible, radial progression is more stable,
and cross-modal organization is less distorted by weak residual interactions. This does
not imply exact recovery of an underlying semantic tree. Rather, it indicates that RAHA
better preserves relative coarse-to-fine structure across modalities, which is precisely
the geometric behavior that the hyperbolic formulation is intended to induce.

%%%%%%%%%%%%%%%%%%%%%%%%%%%%%%%%%%%%%%%%%%%%%%%%%%%%%%%%%%%%%%%%%
\subsection{Hyperparameter Sensitivity and Ablations}
\label{sec:hyperparameter_appendix}
\begin{figure}[t]
\centering
\begin{minipage}[t]{0.24\columnwidth}
\centering
\includegraphics[width=\linewidth]{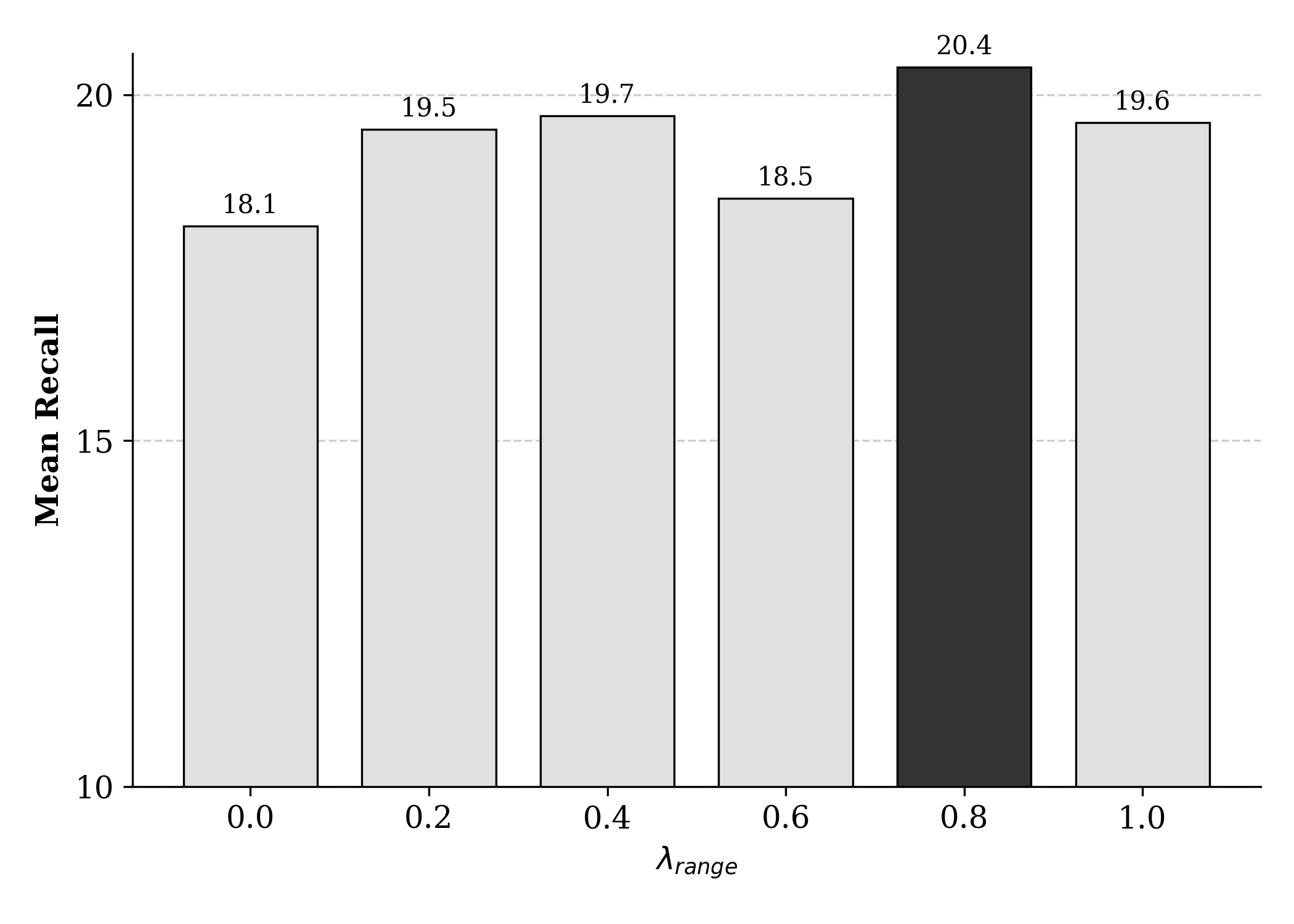}
\vspace{0.2em}
\scriptsize (a) $\lambda_{\mathrm{range}}$
\end{minipage}\hfill
\begin{minipage}[t]{0.24\columnwidth}
\centering
\includegraphics[width=\linewidth]{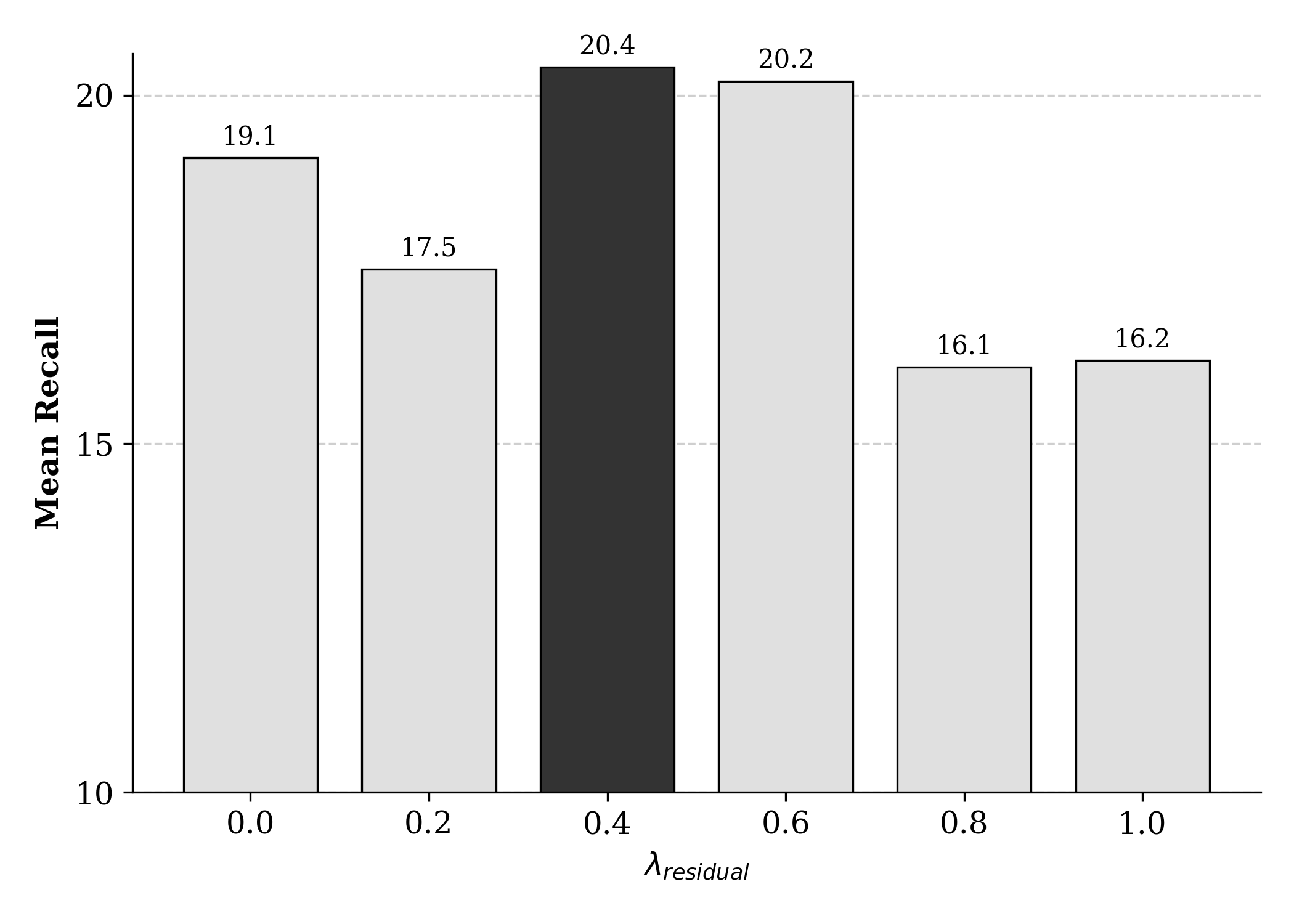}
\vspace{0.2em}
\scriptsize (b) $\lambda_{\mathrm{residual}}$
\end{minipage}\hfill
\begin{minipage}[t]{0.24\columnwidth}
\centering
\includegraphics[width=\linewidth]{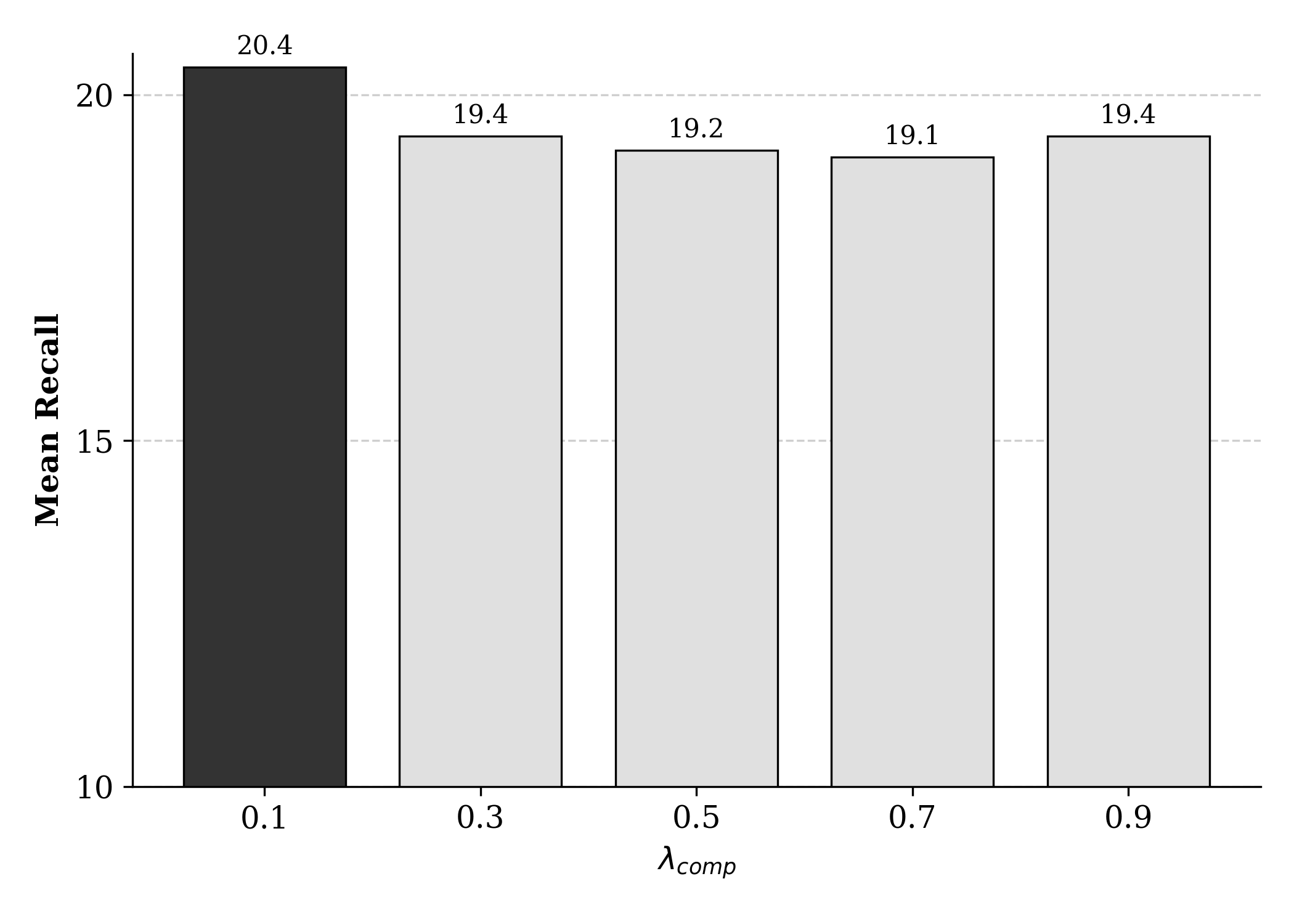}
\vspace{0.2em}
\scriptsize (c) $\lambda_{\mathrm{comp}}$
\end{minipage}\hfill
\begin{minipage}[t]{0.24\columnwidth}
\centering
\includegraphics[width=\linewidth,trim={0 1cm 0 0},clip]{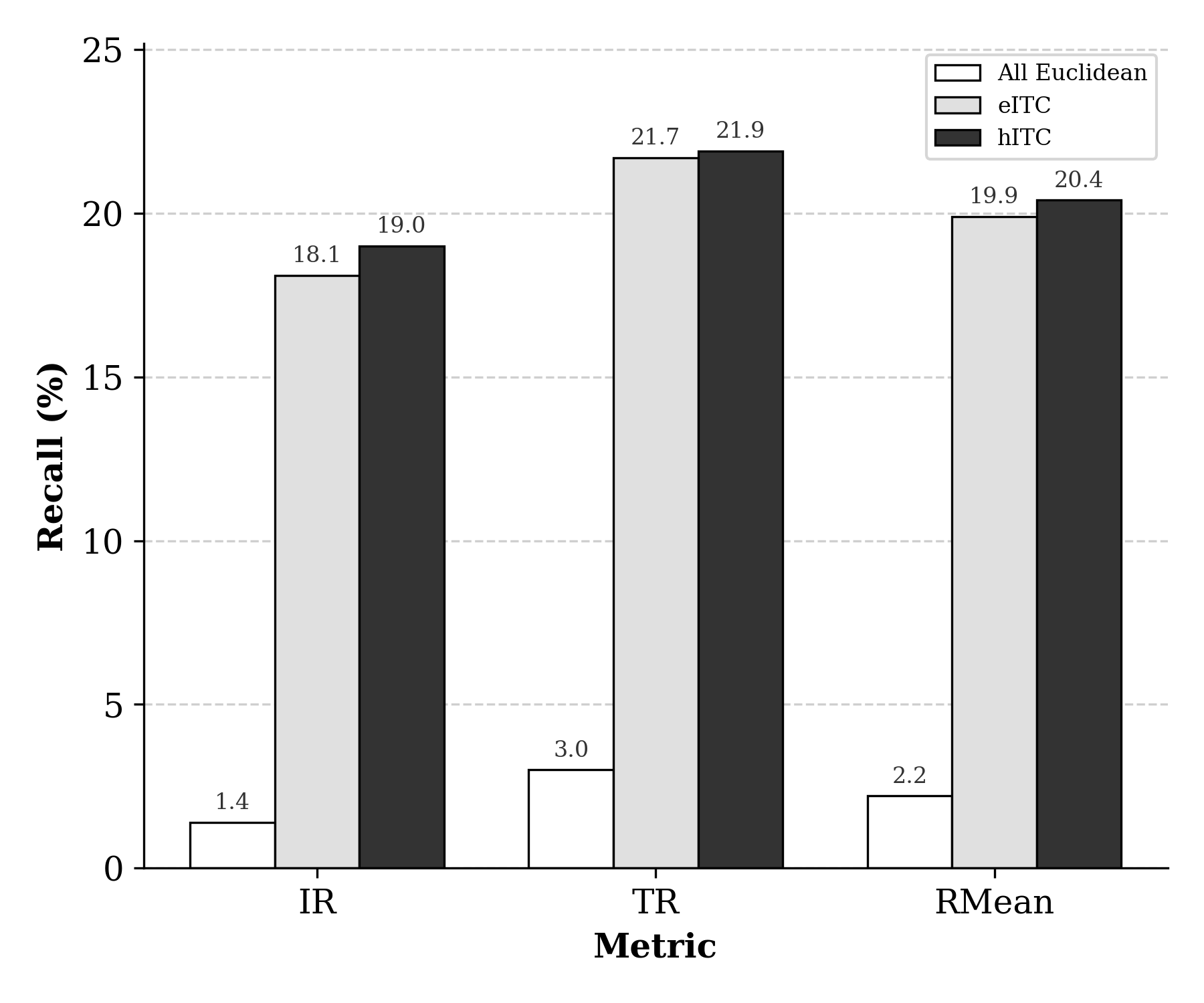}
\vspace{0.2em}
\scriptsize (d) Euclidean vs.\ Hyperbolic space
\end{minipage}
\caption{Hyperparameter sensitivity and geometry ablation of RAHA at the
Flickr8k 100-pair setting. (a)--(c) RAHA is most stable when the dominant shared
range remains the primary alignment target, residual matching is weighted
moderately, and residual compression is kept mild. (d) Keeping the same
Sinkhorn-based relevance-matching pipeline and varying only the geometry shows
that an all-Euclidean formulation performs poorly
(IR/TR/RMean $= 1.4/3.0/2.2$), a mixed variant with Euclidean ITC and
hyperbolic relevance matching (\texttt{eITC}) reaches $18.1/21.7/19.9$, and the
fully hyperbolic variant (\texttt{hITC}) performs best at $19.0/21.9/20.4$.}
\label{fig:budget_diagnostics}
\end{figure}

Taken together with Fig.~{\textcolor{red}{2}} in the main paper, Fig.~\ref{fig:budget_diagnostics} provides a component-wise
view of where RAHA’s gain comes from. Fig.~2 isolates the contribution of the
range--residual decomposition: starting from the base hyperbolic contrastive
objective $\mathcal{L}_{\mathrm{hITC}}$, adding the range term
$\mathcal{L}_{\mathrm{range}}$ yields the larger single-component improvement,
whereas using the residual term $\mathcal{L}_{\mathrm{residual}}$ alone is weaker.
The full objective nevertheless performs best, indicating that the dominant
shared range carries the primary retrieval signal, while the residual branch is
most useful as a controlled secondary refinement once it is anchored to the range
term and regularized by compression. Fig.~\ref{fig:budget_diagnostics}(a)--(c) supports the same
interpretation quantitatively, where stronger range supervision is beneficial for the range branch as opposed to and mild residual compression being
sufficient for the residual branch.

Fig.~\ref{fig:budget_diagnostics}(d) further isolates the role of geometry while keeping the same
Sinkhorn-based relevance-matching pipeline. An all-Euclidean formulation
performs poorly (IR/TR/RMean $= 1.4/3.0/2.2$), showing that the explicit
subspace decomposition is not effective when both contrastive alignment and
relevance matching remain in Euclidean space. Replacing only the contrastive
term by its Euclidean counterpart while keeping the rank-aware relevance
matching in the hyperbolic pipeline (\texttt{eITC}) recovers most of the
performance (IR/TR/RMean $= 18.1/21.7/19.9$), and the fully hyperbolic variant
(\texttt{hITC}) further improves to $19.0/21.9/20.4$. Thus, the gain of RAHA is
not attributable to a single scalar weight or a single loss term in isolation.
It comes from the interaction between selective range--residual supervision and
hyperbolic lifting: the latter is necessary for the decomposition to become
effective, while lifting the contrastive term as well provides an additional
consistent gain over the mixed-geometry variant.

%%%%%%%%%%%%%%%%%%%%%%%%%%%%%%%%%%%%%%%%%%%%%%%%%%%%%%%%%%%%%%%%%
\subsection{Compute Profile}
\label{sec:cost_analysis_appendix}

RAHA follows the line of distribution matching branch for VLDD and avoids the storage overhead of expert-trajectory approaches, which require full parameter checkpoints at each expert step.
At the same time, each distillation step is more expensive than CovMatch because it includes hyperbolic lifting and log-map operations ($\mathcal{O}(Bd)$ element-wise operations), tangent-space cross-covariance and SVD ($\mathcal{O}(d^2 B + d^3)$), range--residual projection ($\mathcal{O}(Bdk)$), and Sinkhorn iterations ($\mathcal{O}(n_{\mathrm{iter}} \cdot \tilde{B} \cdot B)$).

Table~\ref{tab:cost} reports a per-component breakdown on a single RTX A6000 under the default Flickr8k $N{=}100$ setting with batch size~64. All timing values for the distillation step are reported \emph{per synthetic image} within a batch of 64, and the per-iteration wall-clock time is obtained by multiplying by the synthetic batch size~$\tilde{B}$.

\begin{table}[h]
\centering
\renewcommand{\arraystretch}{1.15}
\setlength{\tabcolsep}{4pt}
\caption{Per-component cost breakdown (Flickr8k 100-pair setting on $1{\times}$ RTX A6000). Distillation-step timings are reported per synthetic image at batch size~64.}
\label{tab:cost}
\vspace{-3mm}
\begin{tabular}{lcc}
\toprule
\textbf{Stage} & \textbf{CovMatch}~\cite{lee2025covmatch} & \textbf{RAHA} (Ours) \\
\midrule
Data initialization (per run) & \multicolumn{2}{c}{$\approx$0.69\,s} \\
\midrule
Model initialization (per iter) & \multicolumn{2}{c}{$\approx$0.27\,s}\\
\midrule
Distillation step (per image, $\tilde{B}{=}64$) & $\approx$0.78\,s & $\approx$7.42\,s  \\
Distillation step (per iter, $\tilde{B}{=}64$) & $\approx$55\,s & $\approx$400\,s  \\
\midrule
Peak GPU VRAM & \multicolumn{2}{c}{$\approx$9.3\,GB}\\
\bottomrule
\end{tabular}
\end{table}

The distillation-step overhead is dominated by the SVD computation on the $d{\times}d$ cross-covariance (with $d{=}2304$) and the Sinkhorn iterations on the $\tilde{B}{\times}B$ cost matrix.
At batch size~1, the per-image distillation times are comparable ($\approx$25\,s each), confirming that the overhead scales with batch-level matrix operations rather than per-sample computation.
Peak GPU memory is identical because the additional intermediates (cross-covariance, SVD factors, coupling matrix) are small relative to the model and batch tensors stored by both methods.

Under the default setting of $T{=}200$ iterations with $N_{\mathrm{out}}{=}50$ outer-loop steps each, a full RAHA distillation run on Flickr8k $N{=}100$ completes in approximately 1.3 GPU-hours on a single RTX A6000, compared to approximately 0.14 GPU-hours for CovMatch.
%
% \noindent\textbf{Cost analysis.}\quad
For each distillation iteration, both CovMatch and RAHA occupy a peak GPU memory of approximately 9.3\,GB with batch sizes of 64 for real and synthetic data.  A detailed per-component cost breakdown with contextualization is provided Table~\ref{tab:cost}.

\noindent\textbf{Contextualizing the overhead.}
Three considerations place this cost in practical perspective.
\emph{(i)}~Distillation is a \emph{one-time offline cost}: the resulting synthetic set is reused across all downstream training runs, architecture searches, and ablations.  The amortized cost per downstream experiment is therefore negligible once the distilled set is produced.
\emph{(ii)}~RAHA avoids the storage burden of trajectory-matching methods (e.g., MTT-VL~\cite{wu2024vldistill}, LoRS~\cite{xu2024lors}), which must checkpoint full encoder parameters at every expert step.  For a model with ${\sim}$90\,M parameters saved at 50 expert steps, this amounts to ${\sim}$18\,GB of storage per distillation run---a cost that RAHA does not incur.
\emph{(iii)}~The overhead is concentrated in batch-level matrix operations (SVD, Sinkhorn) rather than per-sample computation.  As shown in the main paper (\S4.4), at batch size~1 both methods take $\approx$25\,s per image; the gap emerges because RAHA's structure-aware operations scale with batch size while CovMatch's element-wise covariance matching does not.

Taken together, these factors indicate that RAHA's compute profile is practical for a distillation method.  We note that the compute overhead is \textbf{not fundamental to the hyperbolic formulation} for VLDD,
% ---the exponential and logarithmic map operations add only $\mathcal{O}(Bd)$ element-wise transcendental functions---
but \textbf{rather to the explicit subspace decomposition and permutation-invariant matching} that constitute the core algorithmic contribution.  As shown in Fig.~\ref{fig:budget_diagnostics}(d), this explicit subspace decomposition does not perform well in Euclidean space, which underscores the importance of the hyperbolic lifting.

%%%%%%%%%%%%%%%%%%%%%%%%%%%%%%%%%%%%%%%%%%%%%%%%%%%%%%%%%%%%%%%%%
\section{Broader Societal Impact}
\label{sec:impact_appendix}

\noindent\fbox{%
\parbox{\dimexpr%
    \linewidth-2\fboxsep-2\fboxrule}%
    {In this section, we discuss the societal implications of image--text dataset distillation, including benefits, inherited risks, and RAHA-specific considerations regarding bias in the preserved subspace.}
}
\vspace{1mm}

Image--text dataset distillation can have both positive and negative downstream effects. On the positive side, compact distilled sets reduce storage, transfer, and repeated training cost. This can make experimentation more accessible and improve reproducibility. Distillation may also reduce the need to redistribute raw paired datasets in settings where privacy, licensing, or provenance concerns limit direct sharing of the original data.

At the same time, distillation does not remove issues present in the source data. A compact synthetic set can still encode harmful correlations, stereotypes, geographic imbalance, or annotation artifacts inherited from the original collection. In multimodal settings, these risks may be amplified by the interaction between image and language. Distillation therefore changes the form of the data, but not automatically its fairness or safety properties.

A consideration specific to RAHA is that the rank-aware decomposition partitions cross-modal correlation into a dominant range subspace and a complementary residual component, with the distillation objective prioritizing preservation of the former.
If harmful correlations, such as stereotypical associations between visual attributes and textual captions, are concentrated in the dominant singular directions, the range-preserving objective will retain them in the distilled set.
Conversely, if such correlations reside primarily in the residual subspace, the compression regularizer may suppress them, but this suppression is incidental rather than by design.
In neither case does RAHA provide an explicit mechanism for bias detection or removal.
We view this as an important direction for future work: combining structure-aware distillation with explicit fairness constraints or post-hoc auditing of the preserved subspace could yield distilled sets that are both efficient and more equitable.

RAHA also has technical limitations that matter for responsible use. As with prior multimodal distillation methods, performance depends on the pretrained encoders used during distillation. It can degrade under strong domain shift, noisy captions, or abstract descriptions that do not exhibit stable hierarchical structure. In addition, the rank-aware prior is most suitable when the shared cross-modal signal is concentrated in a dominant subspace. If useful signal is distributed across many weak directions, excessive residual suppression can remove task-relevant information.

The cost overhead is derived primarily from lifting features to hyperbolic space and taking the tangent, alongside performing SVD on the $d{\times}d$ cross-covariance matrix. and Sinkhorn on the $\tilde B{\times}B$ cost matrix, not from the Lorentz lift. This is a one-time offline distillation cost, and the distilled set can be reused for downstream training. RAHA also avoids trajectory-checkpoint storage required by \cite{wu2024vldistill, xu2024lors, zhang2025repblend}. Thus, \cite{lee2025covmatch} is preferable when wall-clock efficiency is primary, while RAHA is justified when structured relevance modeling, transfer, and higher-budget scaling are prioritized. These bottlenecks are optimizable through truncated/randomized SVD, cached basis updates, low-rank or warm-start Sinkhorn, fewer iters, and fused GPU kernels.

For these reasons, we view RAHA as a method for efficient and structured compression, not as a guarantee of fairness, neutrality, or robustness beyond the tested setting. Future work should study whether structure-aware distillation can be combined with explicit bias auditing, provenance constraints, and safer data filtering.

\bibliographystyle{splncs04}
\bibliography{main}
\end{document}